# Evaluating and Explaining Earthquake-Induced Liquefaction Potential through Multi-Modal Transformers

Sompote Youwai[1], Tipok Kitkobsin, Sutat Leelataviwat and Pornkasem Jongpradist

AI Research Group
Department of Civil Engineering
Faculty of Engineering
King Mongkut's University of Technology Thonburi

[1]Corresponding author

**Abstract**

This study presents an explainable parallel transformer architecture for soil liquefaction prediction that integrates three distinct data streams: spectral seismic encoding, soil stratigraphy tokenization, and site-specific features. The architecture processes data from 165 case histories across 11 major earthquakes, employing Fast Fourier Transform for seismic waveform encoding and principles from large language models for soil layer tokenization. Interpretability is achieved through SHapley Additive exPlanations (SHAP), which decompose predictions into individual contributions from seismic characteristics, soil properties, and site conditions. The model achieves 93.75% prediction accuracy on cross-regional validation sets and demonstrates robust performance through sensitivity analysis of ground motion intensity and soil resistance parameters. Notably, validation against previously unseen ground motion data from the 2024 Noto Peninsula earthquake confirms the model's generalization capabilities and practical utility. Implementation as a publicly accessible web application enables rapid assessment of multiple sites simultaneously. This approach establishes a new framework in geotechnical deep learning where sophisticated multi-modal analysis meets practical engineering requirements through quantitative interpretation and accessible deployment.

## 1. Introduction

Soil liquefaction during seismic events represents one of the most critical phenomena in geotechnical engineering, posing substantial risks to infrastructure integrity and human safety. The complex, non-linear interactions between seismic waves and spatially variable soil properties make the accurate prediction of liquefaction susceptibility both crucial for civil infrastructure resilience and inherently challenging. Traditional methodologies, primarily founded on empirical correlations and simplified analytical frameworks (Andrus and Stokoe Ii, 2000; Boulanger and Idriss, 2016; Cetin et al., 2018; Robertson and Wride, 1998; Seed and Idriss, 1971), often demonstrate limited capacity to capture the multifaceted nature of soil-earthquake interactions across diverse geological contexts. A significant limitation of conventional approaches lies in their discrete analysis methodology, which typically evaluates liquefaction potential at isolated soil depths rather than considering the holistic response of the entire soil profile. Furthermore, these methods often reduce complex seismic events to simplified parameters, such as single-value magnitude representations, thereby neglecting the temporal and spectral characteristics of ground motion. This reductionist approach may introduce substantial uncertainties in liquefaction susceptibility assessments, potentially compromising the reliability of prediction models.

Artificial neural networks (ANNs) have emerged as the dominant methodology in AI-based soil liquefaction prediction, accounting for 77% of published studies according to a comprehensive review by (Maurer and Sanger, 2023). Despite their widespread adoption (Abbaszadeh Shahri, 2016; Fahim et al., 2022; Goh, 1996; Hanna et al., 2007; Hsein Juang et al., 1999; Kurup and Garg, 2005), ANNs present notable limitations: they require extensive training datasets and operate as "black boxes," making their decision-making processes opaque to practitioners. While alternative approaches such as decision trees (Demir and Sahin, 2022) and support vector machines (Goh and Goh, 2007) have been explored, their application remains limited, suggesting a tendency to default to popular architectures rather than identifying optimal solutions for specific liquefaction challenges. To address these interpretability challenges, Explainable Artificial Intelligence (XAI) has emerged as a crucial advancement in modern AI systems. While AI has demonstrated remarkable success across domains including healthcare (Lai, 2024) and finance (Chen et al., 2023), the opacity of advanced models, particularly deep learning architectures, has historically limited their practical adoption in geotechnical engineering. XAI methodologies have proven valuable across various geotechnical applications, including slope stability analysis (Abdollahi et al., 2024), tunnel assessment (Liu et al., 2024), and liquefaction prediction (Hsiao et al., 2024). Our research integrates XAI with Transformer methodology for liquefaction potential evaluation, employing both local and global interpretability frameworks to decode feature influences on liquefaction susceptibility.

This integrated approach reveals two critical limitations in current practice. First, existing models typically analyze discrete soil parameters at isolated depths, overlooking the spatial correlations and continuous distributions of soil properties across both micro and macro scales. A more effective approach would consider the entire soil profile to better capture site-specific geotechnical behavior. Second, current frameworks inadequately address seismic excitation factors, particularly lacking comprehensive integration of ground motion characteristics in time and frequency domains. The fundamental parameters of earthquake magnitude and duration, crucial for liquefaction initiation, demand more sophisticated treatment within AI prediction models. Nevertheless, XAI techniques effectively illuminate feature importance patterns in both accurate and inaccurate predictions, providing valuable insights across local and global scales. This enhanced interpretability bridges the gap between advanced AI architectures and geotechnical practitioners, facilitating the adoption of these sophisticated models in practical engineering applications.

This research advances the transformer-based architecture proposed by (Youwai and Detcheewa, 2025) through a novel dual-stream approach to soil liquefaction prediction (Fig. 1). The architecture processes two distinct data streams: soil profiles and seismic ground motions. The first stream handles soil profiles derived from field tests (standard penetration and cone penetration tests) collected from 163 global locations. These depth-specific characteristics are tokenized, analogous to language model tokenization, and projected into an embedded vector space with positional encodings to preserve depth relationships. A transformer encoder then processes these embedded tensors, capturing the spatial correlations within soil profiles. The second stream analyzes earthquake ground motion data by encoding frequency-domain transformations of temporal signals through a parallel transformer network, enabling comprehensive analysis of seismic characteristics. The outputs from both streams converge in a multilayer perceptron for final liquefaction potential assessment. The model's robustness was validated using a comprehensive database spanning 11 major earthquakes across diverse geological settings. The architecture's innovative integration of frequency-domain analysis with hierarchical attention mechanisms enables sophisticated interpretation of both temporal and spatial features in liquefaction prediction. Model interpretability is achieved through SHAP (SHapley Additive exPlanations) values, providing both global feature importance analysis and local explanations for specific predictions. This dual-level interpretability framework enables practitioners to understand key factors influencing liquefaction potential at both regional and site-specific scales.

The developed framework yields statistically significant performance gains compared to conventional methods while preserving complete transparency in its decision-making processes. Through the integration of state-of-the-art deep learning architectures with comprehensive interpretability frameworks, this approach establishes a novel paradigm in geotechnical engineering that reconciles advanced predictive capabilities with essential interpretability requirements. To enable broad implementation across the field, we have deployed the model and associated interpretability tools via a web-based interface designed for civil engineers without computational expertise. The manuscript is structured as follows: Section 2 details the model architecture; Section 3 presents the experimental methodology and results; Section 4 elucidates the explainable artificial intelligence framework; Section 5 demonstrates the application to an industry case study; Section 6 examines the implications and limitations of the approach; and Section 7 concludes with principal findings and future research directions.

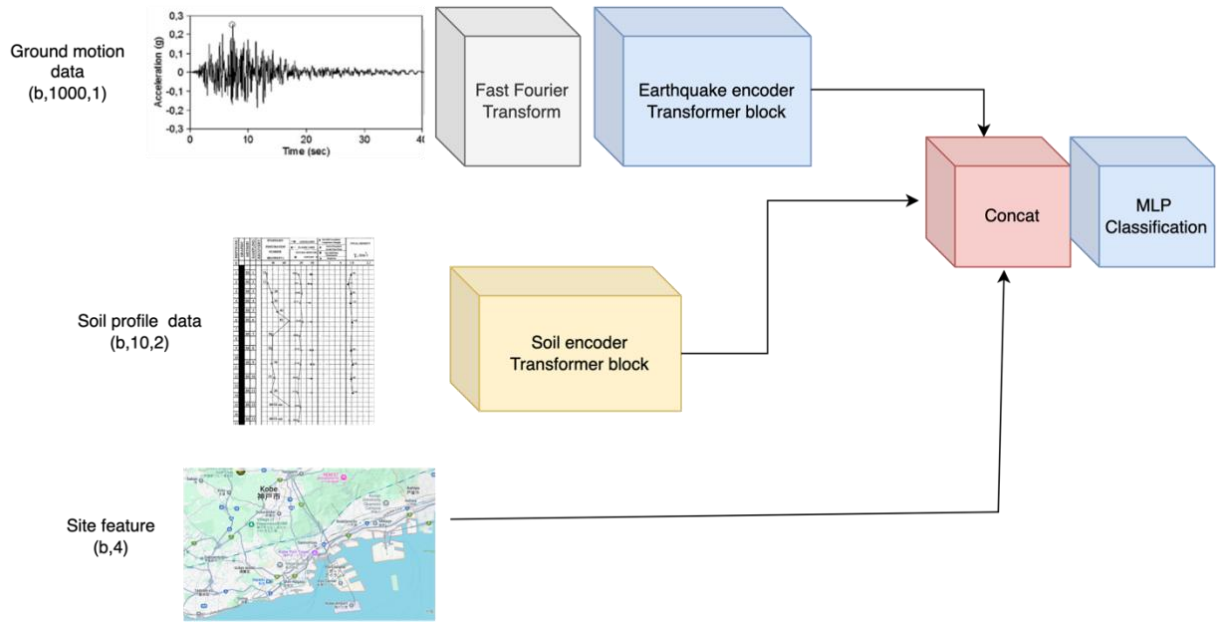

Figure 1 Architecture of the model

## 2. Model architecture

The proposed model employs a tripartite architecture comprising three distinct streams for comprehensive liquefaction potential assessment (Figs. 1 and 2). The first stream focuses on ground motion encoding, where temporal acceleration data undergoes Fast Fourier Transform (FFT) to capture frequency-domain characteristics before being processed through a transformer block. The second stream incorporates a soil profile encoder, where sequential data including soil strength parameters and soil type classifications are processed through a transformer architecture, preserving the spatial relationships between soil layers. The third stream processes site-specific features, including groundwater proximity, 30-meter shear wave velocity (Vs30), epicentral distance, and groundwater level. These three components are concatenated and fed into a multilayer perceptron for the final classification of liquefaction potential at the site of interest.

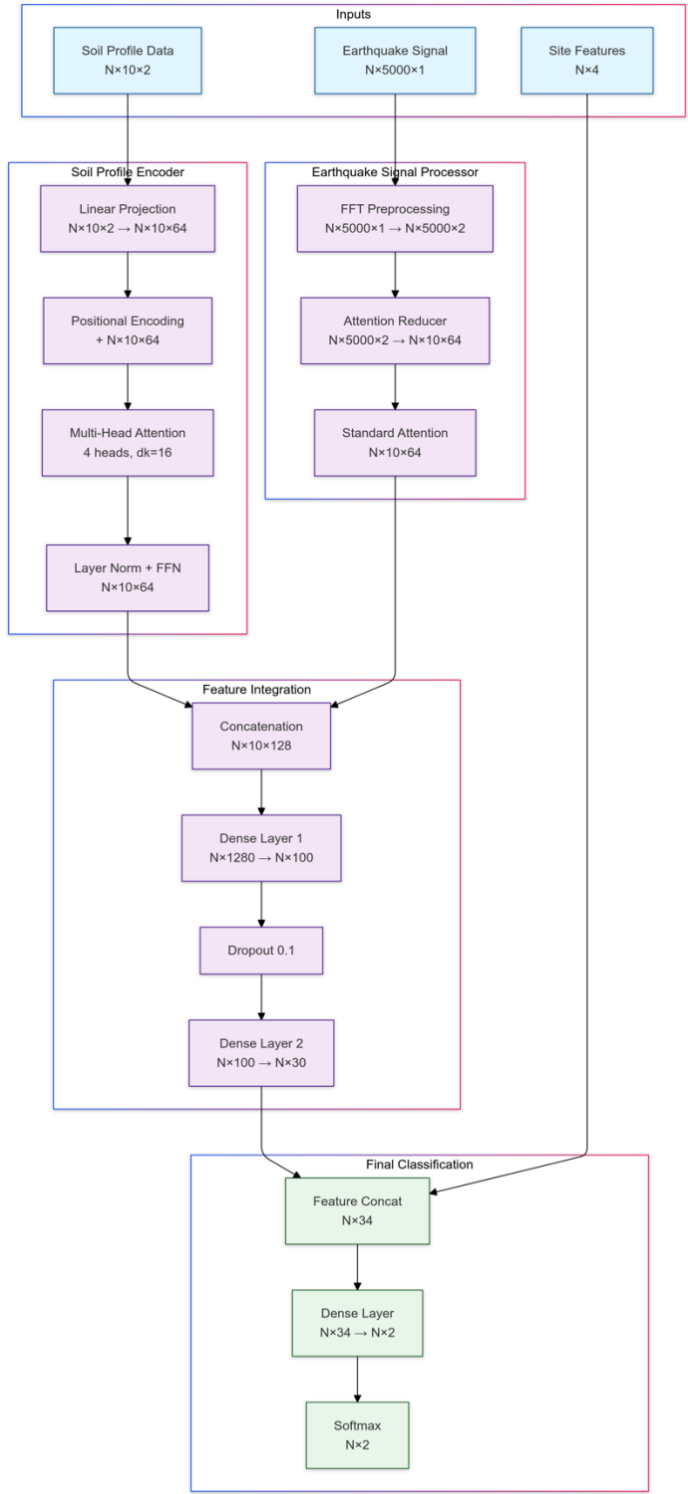

Figure. 2 Model architecture detail

The liquefaction susceptibility assessment framework incorporates diverse geotechnical parameters through a comprehensive mathematical formulation, illustrated in Figure 2. The input data architecture comprises three fundamental components. The soil characteristics are quantified via Standard Penetration Test (SPT) measurements and soil classification matrices, denoted as $X_{SPT} \in R^{n \times d}$ and $X_{soiltype} \in R^{n \times d}$, respectively, where n represents the number of sampling points and d denotes the depth intervals. $X_{SPT}$ contains the SPT-N values at discrete depths, while $X_{soil}$ represents a categorical encoding of soil types, where {1, 2, 3} correspond to sand, silty sand, and clay, respectively. These matrices are subsequently concatenated to form the composite input tensor

$$X_{soil} = [X_{SPT} || X_{soiltype}] \in R^{n \times d \times 2}. \tag{1}$$

The soil input tensor then undergoes linear projection from the input space to a higher-dimensional embedded space $E \in R^{n \times d \times h}$, where h represents the embedding dimension. This projection serves three crucial purposes: (1) it maps the heterogeneous input features into a uniform latent space, enabling the transformer to process both continuous SPT values and categorical soil types consistently; (2) it increases the dimensionality of the feature space, allowing the model to capture more complex patterns and interactions between soil parameters; and (3) it creates learnable embeddings that can adapt during training to better represent the relationships between different soil characteristics and their contribution to liquefaction susceptibility. Second, earthquake data is captured as a time series representation $X_{eq} \in R^{N \times T \times 1}$. Third, site features including location and additional parameters are represented as $X_{site} \in R^{N \times F}$. Data preprocessing plays a crucial role in the model's effectiveness. Numerical features undergo standardization using the formula

$$X_{standardized} = \frac{X - \mu}{\sigma}. \tag{2}$$

The model's architecture encompasses specific dimensions for various layers. Input layers are defined as:

$$X_{soil} \in R^{N \times 10 \times 2} \tag{3}$$
$$X_{eq} \in R^{N \times T \times 1} \tag{4}$$
$$X_{site} \in R^{N \times 4} \tag{5}$$

Intermediate representations maintain the following dimensions:

$$H_{soil} = f_{soil}(X_{soil}) \tag{6}$$
$$H_{eq} = f_{eq}(X_{eq}) \tag{7}$$

$$H_{soil} \in R^{N \times 10 \times 64} \tag{8}$$
$$H_{eq} \in R^{N \times 10 \times 64} \tag{9}$$
$$H_{combined} \in R^{N \times 10 * 128} \tag{10}$$

The complete forward pass can be expressed as a composition of functions:

The integration network combines and processes features through multiple layers:
$$H_{combined} = Concat[H_{soil}; H_{eq}] \tag{11}$$
$$H_1 = LeakyReLU(W_1 H_{combined} + b_1) \tag{12}$$
$$H_2 = LeakyReLU(W_2 Concat[H_1; X_{site}] + b_2) \tag{13}$$

$$Y_{pred} = \text{softmax}(W_3 H_2 + b_3) \tag{14}$$

The output layer produces predictions in the form $Y_{pred} \in R^{N \times 2}$

## 2.1 Earthquake signal processor

Fourier transformation to capture its frequency-domain characteristics. Given a time-domain acceleration signal $x(t)$, the Fast Fourier Transform (FFT) is applied to obtain the frequency spectrum $X(f)$:

$$X(f) = \mathcal{F}\{x(t)\} = \int_{-\infty}^{\infty} x(t) e^{-2\pi i f t} \, dt \tag{16}$$

The resulting complex-valued frequency spectrum is then normalized to ensure numerical stability and consistent scale across different seismic events. The normalized spectrum $X_{norm}(f)$ is computed as:

$$X_{norm}(f) = \frac{|X(f)|}{\sqrt{\sum_f |X(f)|^2 + \epsilon}} \tag{17}$$

where $X(f)$ represents the magnitude spectrum and $\epsilon$ is a small positive constant (typically $10^{-8}$) added to prevent division by zero. This normalization step ensures that the frequency components are scaled relative to the total spectral energy while maintaining numerical stability in the subsequent transformer processing stages.

1. Input Projection and Positional Encoding:
The input tensor $X \in R^{B \times L \times 2}$ is first projected to a higher dimension:

$$X_{proj} = X W_{proj} + b_{proj}, \quad W_{proj} \in R^{2 \times 64} \tag{18}$$

Then positional encoding $P$ is added to incorporate sequence order information:
$$X_{pos} = X_{proj} + P \tag{19}$$

where P uses sinusoidal encodings:
$$P_{pos,2i} = \sin(pos/10000^{2i/d_{model}}) \tag{20}$$
$$P_{pos,2i+1} = \cos(pos/10000^{2i/d_{model}}) \tag{21}$$

2. Multi-Head Self-Attention:

For each head $h \in \{1, \ldots, H\}$, the input is projected into queries, keys, and values:

$$Q_h = X_{pos} W_h^Q, \quad K_h = X_{pos} W_h^K, \quad V_h = X_{pos} W_h^V \tag{22}$$

where $W_h^Q, W_h^K, W_h^V \in R^{64 \times d_h}$ and $d_h = 64/H$ is the head dimension.

The attention scores for each head are computed as:

$$Z_h = \text{softmax}\left(\frac{Q_h K_h^T}{\sqrt{d_h}}\right) V_h \tag{23}$$

The softmax operation normalizes attention scores across the key dimension:

$$\text{softmax}(x_i) = \frac{\exp(x_i)}{\sum_j \exp(x_j)} \tag{24}$$

3. Residual Connection and Layer Normalization:
The attention outputs are concatenated and projected (Head=2):

$$\text{MultiHead}(X) = \text{Concat}(Z_1, \ldots, Z_H) W^O \tag{25}$$

A residual connection and layer normalization are applied:

$$X_{out} = \text{LayerNorm}\left(X_{pos} + \text{MultiHead}(X_{pos})\right) \tag{26}$$

Layer normalization computes:

$$\text{LayerNorm}(x) = \gamma \odot \frac{x - \mu}{\sqrt{\sigma^2 + \epsilon}} + \beta \tag{27}$$

Where μ and σ are mean and standard deviation computed across the feature dimension.

4. Dimensionality and Sequence Length Reduction:
The output is processed through two fully connected layers:

$$X_{fc} = \text{FC1}\left(\text{FC}_{out}(X_{out})\right) \tag{28}$$

Finally, adaptive average pooling reduces the sequence length to $L_{out}$:

$$X_{final} = \text{AdaptiveAvgPool1d}(X_{fc}) \tag{29}$$

The pooling operation can be expressed as:

$$X_{final}[i] = \frac{1}{r}\sum_{j=ri}^{r(i+1)-1} X_{fc}[j] \tag{30}$$

Where $r = L/L_{out}$ is the reduction factor.

The complete transformation pipeline can be summarized as:

$$X \xrightarrow{project} X_{proj} \xrightarrow{+pro} X_{pos} \xrightarrow{attention} X_{out} \xrightarrow{FC} X_{fc} \xrightarrow{pool} X_{final} \tag{31}$$

This architecture effectively reduces both the sequence length and dimensionality while maintaining the ability to capture long-range dependencies through the self-attention mechanism. The multiple attention heads allow the model to focus on different aspects of the input sequence in parallel.

## 2.2 Soil profile encoders

1. Input Projection:
Given input $X \in R^{B \times L \times 2}$:

$$X_{proj} = XW_{proj} + b_{proj}, \quad W_{proj} \in R^{2\times 64} \quad (32)$$

This projection transforms the 2-dimensional input features into a 64-dimensional space, allowing the model to learn richer representations. The bias term $b_{proj}$ adds flexibility to the transformation.

2. Positional Encoding:

$$PE_{(pos,2i)} = \sin(pos/10000^{2i/64}) \quad (33)$$
$$PE_{(pos,2i+1)} = \cos(pos/10000^{2i/64}) \quad (34)$$
$$X_{pos} = X_{proj} + PE \quad (35)$$

Positional encoding adds sequential information to each position in the input sequence. The sinusoidal functions create unique patterns for each position, allowing the model to understand the order of elements. The wavelengths form a geometric progression from $2\pi$ to $10000 \cdot 2\pi$.

3. Multi-Head Attention (4 heads):
For each head $h \in \{1,2,3,4\}$:

$$Q_h = X_{pos}W_h^Q, \quad K_h = X_{pos}W_h^K, \quad V_h = X_{pos}W_h^V \quad (36)$$
$$Z_h = \text{softmax}\left(\frac{Q_h K_h^T}{\sqrt{16}}\right)V_h \quad (37)$$

Each attention head learns different aspects of the relationships between sequence elements. The queries (Q), keys (K), and values (V) are linear projections of the input. The scaling factor $\sqrt{16}$ prevents vanishing gradients in the softmax operation.

Combined attention output:
$$\text{MultiHead}(X) = \text{Concat}(Z_1, \ldots, Z_4)W^O \quad (38)$$

The outputs of all heads are concatenated and projected to create the final attention output, combining different aspects of the learned relationships.

4. Layer Normalization:
$$\text{LayerNorm}(x) = \gamma \odot \frac{x-\mu}{\sqrt{\sigma^2 + \epsilon}} + \beta \quad (39)$$

Layer normalization stabilizes training by normalizing the activations across the feature dimension. The learnable parameters $\gamma$ and $\beta$ allow the model to scale and shift the normalized values as needed.

5. Position-wise Feed-Forward:
For input $x \in R^{B \times L \times d_{model}}$:
$$\text{FFN}(x) = W_2 \text{LeakyReLU}(W_1 x + b_1) + b_2 \quad (40)$$
where:
- $W_1 \in R^{d_{ff} \times d_{model}}$

- $W_2 \in R^{d_{model} \times d_{ff}}$
- LeakyReLU slope = 0.01: $f(x) = \max(0.01x, x)$

This revised FFN uses LeakyReLU instead of standard ReLU, preventing "dying ReLU" problems by allowing small negative gradients.

6. Complete Forward Pass:

$$X_{out} = X_{pos} + \text{LayerNorm}\left(\text{FFN}\left(\text{LayerNorm}\left(X_{pos} + \text{MultiHead}(X_{pos})\right)\right)\right) \quad (41)$$

The entire sequence of transformations with residual connections.

7. Final Processing:

$$X_{final} = X + \text{Dropout}(X_{out}) \quad (42)$$

This architecture leverages LeakyReLU's properties to prevent vanishing gradients in the negative range while maintaining the network's ability to learn non-linear transformations. The position-wise nature means each position in the sequence is transformed independently, preserving positional information while adding model capacity.

## 3. Experiment
### 3.1 Data characteristics

This study utilized the Next Generation Liquefaction (NGL) database to analyze an extensive dataset comprising seismic and geotechnical measurements from 11 major earthquakes (1980-2020), as documented in Figure 3 (Brandenberg et al., 2020). Despite the abundance of global liquefaction case histories in the literature, many existing databases lack comprehensive subsurface soil profiles at documented liquefaction sites, limiting their applicability for method development and validation. This investigation utilized a carefully curated subset of sites from the NGL database with complete Standard Penetration Test (SPT) or Cone Penetration Test (CPT) data, accompanied by detailed soil classifications and groundwater conditions. The final dataset encompassed 165 globally distributed sites across diverse geological settings, with 105 sites (64%) exhibiting liquefaction and 60 sites (36%) showing no evidence of liquefaction. Subsurface investigations at each location extended to minimum depths of 10 meters, facilitating thorough characterization of soil stratigraphy and liquefaction susceptibility. The seismic events primarily occurred along the Pacific Ring of Fire, with epicentral distributions across North America (M 6.5-7.2), Europe (M 6.0-6.8), Asia—specifically Japan and Taiwan (M 6.8-7.6)—and New Zealand (M 6.2-7.1). While the Pacific Ring is typically associated with subduction zone seismicity, this dataset predominantly represents crustal fault events. The earthquakes exhibited thrust and strike-slip faulting mechanisms at hypocentral depths of 10-35 kilometers. The selected case histories from the NGL database, representing diverse geological and seismological conditions, constitute a robust dataset for evaluating liquefaction potential assessment methodologies.

While extensive global liquefaction case histories exist in the literature, a significant limitation in many databases is the absence of detailed subsurface soil profiles at documented liquefaction sites. This constrains the utility of many available case histories for developing or validating liquefaction assessment methods. From the extensive global database, only sites with either Standard Penetration Test (SPT) or Cone Penetration Test (CPT) measurements, along with complete soil classifications and groundwater conditions, were selected for this study. The final dataset comprised 165 global sites distributed across diverse geological settings, with liquefaction observed at 105 locations (64%) and no liquefaction at 60 locations (36%). Each site was characterized by detailed subsurface investigations extending to depths of at least 10 meters, ensuring comprehensive evaluation of soil stratigraphy and liquefaction susceptibility. The earthquakes occurred primarily along the Pacific Ring of Fire, with epicenters distributed across North America (magnitude range: 6.5-7.2), Europe (magnitude range: 6.0-6.8), and Asia—specifically Japan and Taiwan (magnitude range: 6.8-7.6)—as well as New Zealand (magnitude range: 6.2-7.1). The Pacific ring is normally associated with subduction zone earthquakes, but the data here is mostly crustal faults. These seismic events were associated with thrust and strike-slip faulting mechanisms at depths ranging from 10 to 35 kilometers. The selected case histories represent a diverse range of geological and seismological conditions, providing a robust dataset for evaluating liquefaction potential assessment methods.

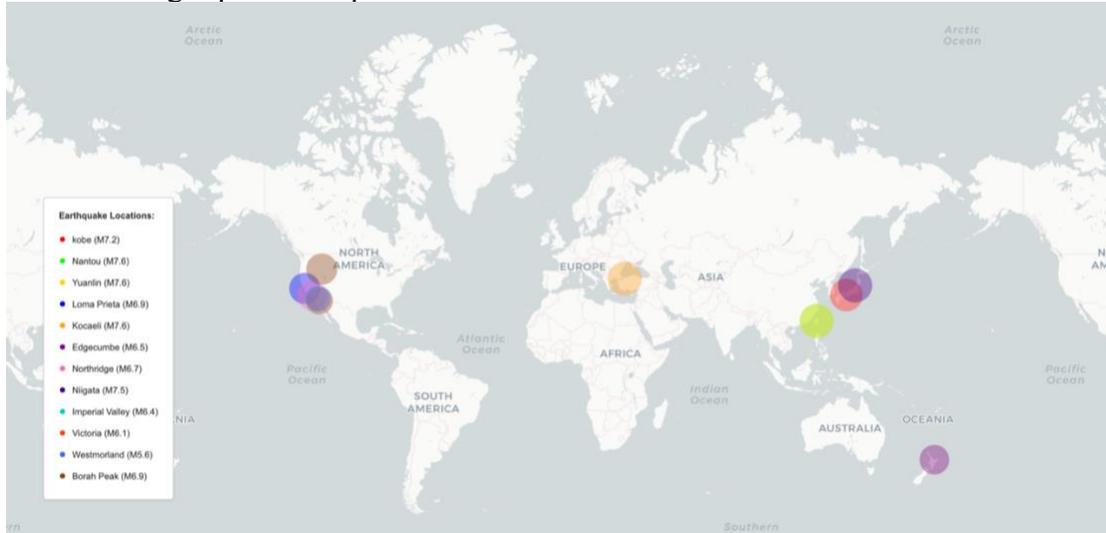

Figure 3 The location of earthquake in this study

The earthquake ground motion records were transformed into the frequency domain using Fast Fourier Transform (FFT) analysis, with results presented in Figure 4, following the methodology described in Section 2.2. Spectral analysis revealed that the Loma Prieta earthquake exhibited the highest amplitude at approximately 2 Hz, followed by the Edgecumbe earthquake in terms of peak spectral amplitude. Notable frequency content characteristics were also observed in the Yualin earthquake, which demonstrated maximum spectral amplitude at 5 Hz. The implementation of frequency domain input parameters was strategically chosen to enable the model to identify and prioritize significant frequency components through adaptive attention weighting during the encoding phase. This approach allows the model to optimize its learning by assigning higher weights to frequencies that correlate strongly with liquefaction occurrence, while automatically attenuating the influence of noise and non-predictive frequency components during the training process. This frequency-based feature selection mechanism enhances the model's ability to distinguish between relevant seismic characteristics and background noise in the ground motion signals.

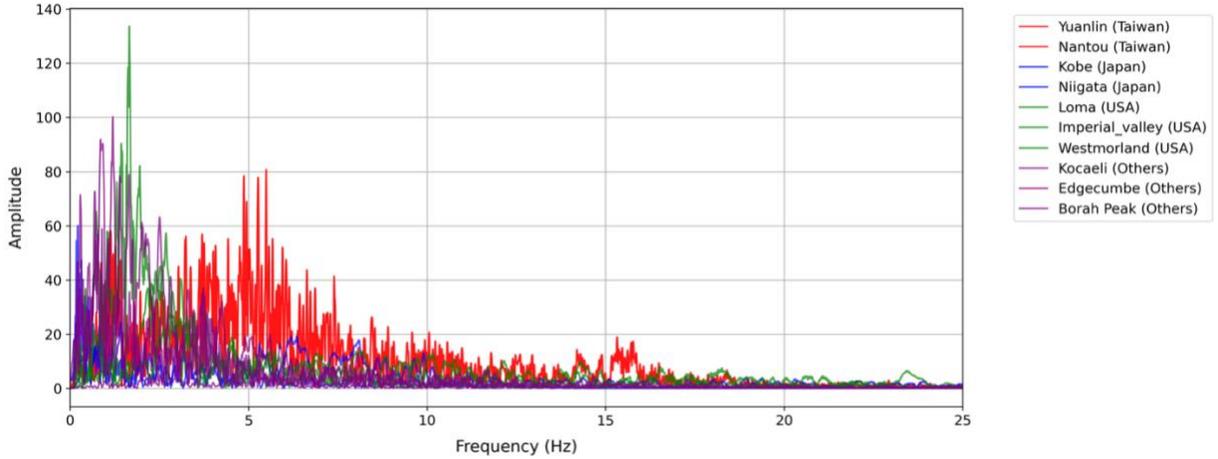

Figure 4 The frequency domain of earthquake used in this study

The analyzed soil profile comprised 165 bore holes, with data sourced from previous case studies in the literature. The model was trained using Standard Penetration Test (SPT) measurements, expressed in blows per foot (N-value). For locations where Cone Penetration Test (CPT) data was available instead of SPT, the values were converted to equivalent SPT measurements using Equation 43 (Bol, 2023).

$$\frac{q_t}{P_a} = AN_{60}^B \qquad (43)$$
$$A = 92.728 \cdot (I_c)^{-2.746} \qquad (44)$$
$$B = -0.1185 \cdot (I_c)^2 + 0.5333 \cdot (I_c) - 0.0764 \qquad (45)$$

where $q_t$ is cone resistance, $P_a$ is the atmospheric pressure (0.1 MPa), $I_c$ for sand = 1.7 for silty sand = 2.2 and for clay = 2.95. The distribution of SPT values, illustrated in Figure 5, demonstrates significant variability across the samples, predominantly indicating loose sand conditions conducive to liquefaction potential. The soil composition analysis, also presented in Figure 5, reveals that soil type significantly influences liquefaction susceptibility. The study investigated three primary soil classifications: sand, silty sand, and cohesive soils (clay and silt). Statistical analysis of the SPT values shows considerable variation within each soil classification, with loose sand deposits exhibiting particularly notable susceptibility to liquefaction. The CPT to SPT conversion methodology followed established correlations from the literature (Bol, 2023), enabling comprehensive integration of both testing methods in the dataset. The soil classification distribution, coupled with corresponding SPT measurements, provided a robust foundation for evaluating liquefaction potential across diverse subsurface conditions.

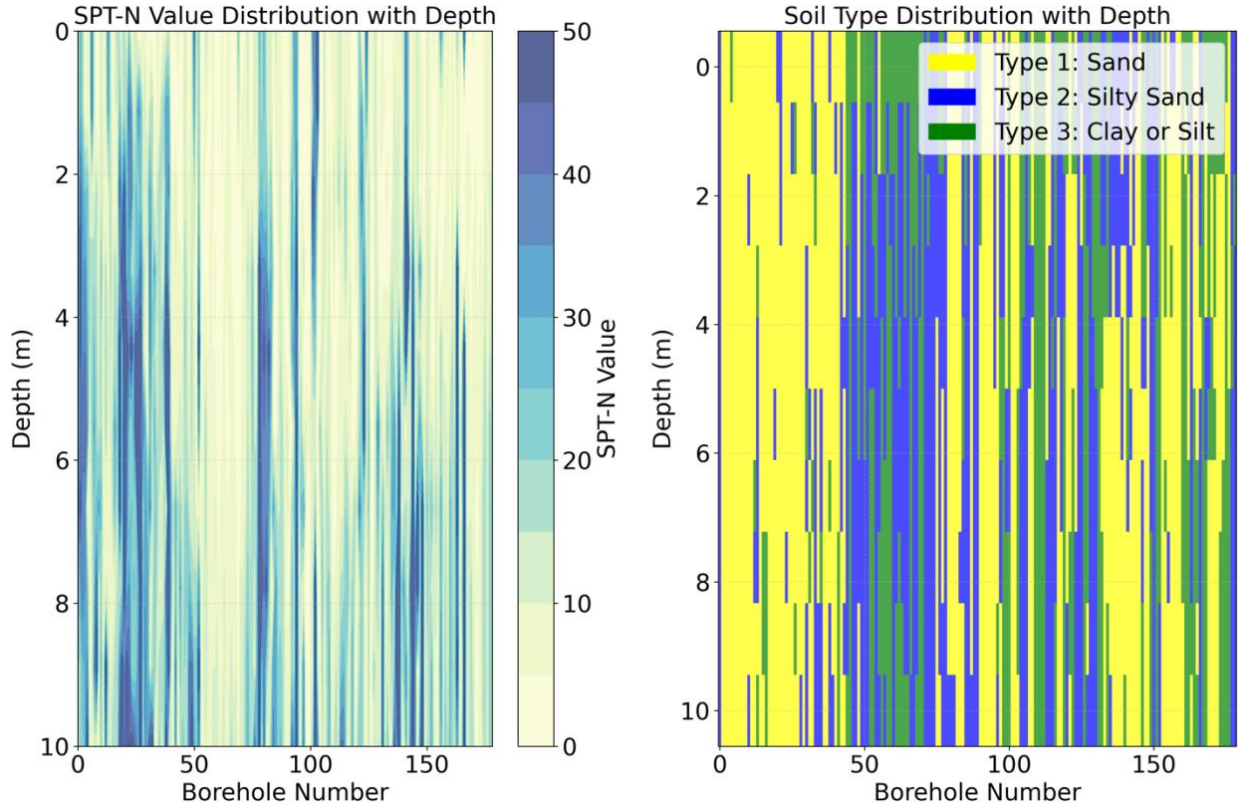

Figure 5 The SPT value and soil type of data

Site characteristics represent critical determinants in earthquake-induced liquefaction potential, significantly influencing the seismic response of soil deposits. The feature tensor incorporated four key parameters: shear wave velocity at 30 m depth (Vs30), epicentral distance, groundwater level, and proximity to water bodies, as illustrated in Figure 6. The Vs30 measurements ranged from 100 to 700 m/s, with predominant values clustering between 200 and 300 m/s, indicating predominantly medium-dense to dense soil conditions. These Vs30 values, obtained from USGS database(USGS, 2025), align with typical ranges associated with liquefaction-susceptible deposits reported in previous seismic hazard assessments The proximity to water bodies, including rivers and lakes, varied from immediately adjacent locations to distances of approximately 35 m, with the majority of sites situated in close proximity to water sources, suggesting potential hydraulic connectivity. Groundwater level, a crucial parameter in liquefaction susceptibility assessment, ranged from ground surface to 10 m below grade across the dataset, with shallow groundwater tables generally associated with increased liquefaction potential. The epicentral distance of the investigated sites exhibited substantial variation, spanning from 1 km to 100 km, providing a comprehensive range of seismic exposure conditions and enabling assessment of spatial attenuation effects on liquefaction triggering. This spatial distribution of parameters enables robust analysis of the relationship between site characteristics and liquefaction potential, while the extensive range of values for each parameter ensures broad applicability of the developed model across diverse geological and seismological settings.

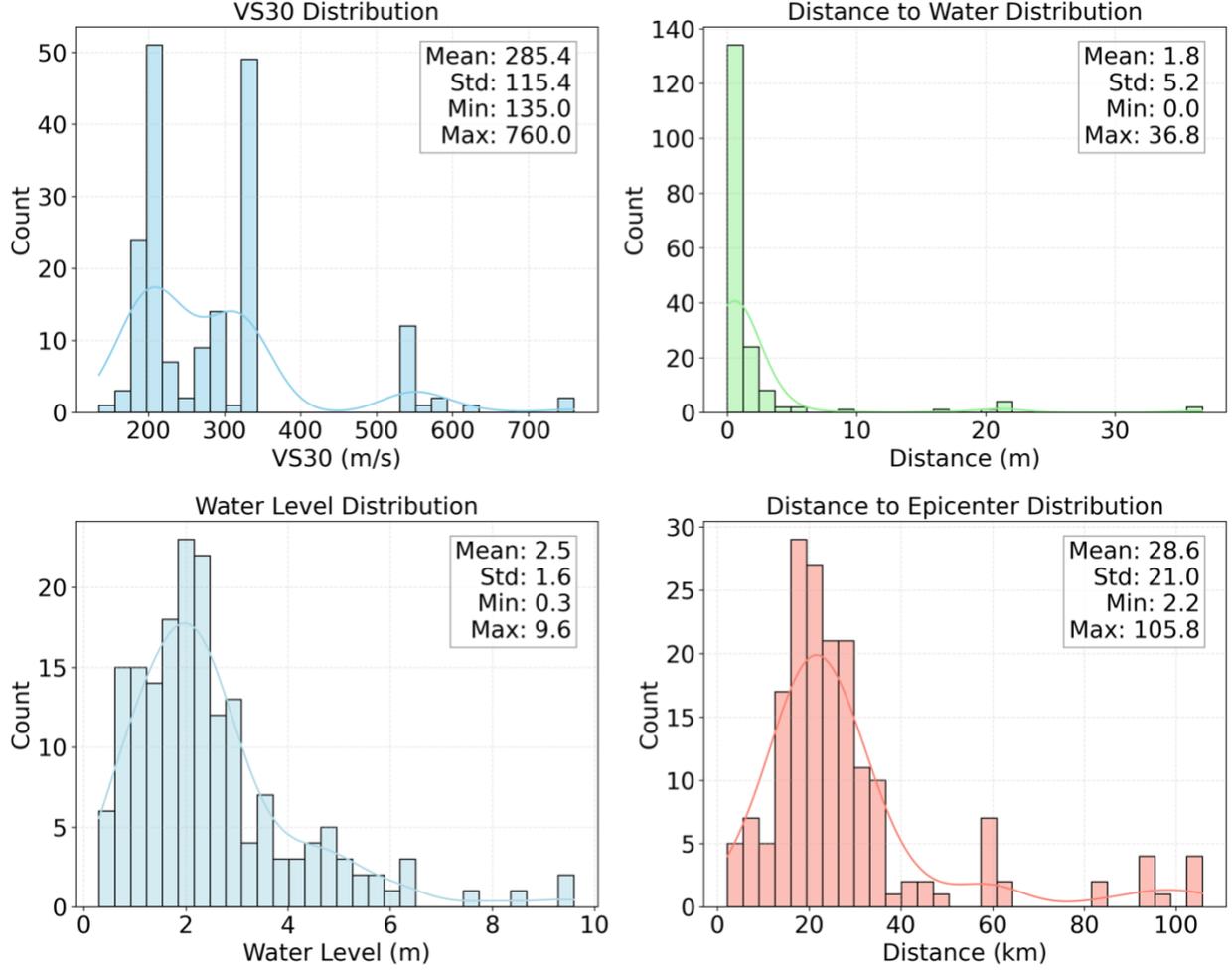

Figure 6 Statistical distributions of seismic site characteristics

## 3.1 Training Methodology

The model's training dataset was augmented using liquefaction case histories from 165 global sites, which originally experienced strong seismic events (moment magnitude Mw ranging from 6.0 to 7.6). To enhance the model's capability to predict liquefaction potential across a broader range of seismic intensities, the dataset was expanded by incorporating null-motion scenarios for each site. This data augmentation technique effectively doubled the training corpus, resulting in 330 data points. The expanded dataset, which included both strong-motion and zero-motion cases, was then utilized for model training and validation. This systematic approach to data augmentation helped address the inherent bias in the original dataset toward high-magnitude events, thereby improving the model's generalization capabilities across various seismic intensity levels. The development of the liquefaction prediction model incorporated robust data preprocessing and training methodologies. The preprocessing phase employed StandardScaler for feature normalization, following the transformation

$$X_{normalized} = \frac{X-\mu}{\sigma} \tag{46}$$

where μ represents the population mean and σ denotes the standard deviation per feature, ensuring equitable feature contribution during model training.

Frequency domain feature extraction utilized Fast Fourier Transform (FFT) preprocessing, computed as

$$X_{fft} = |\mathcal{F}(x)|, \text{ with subsequent magnitude spectrum} \quad (47)$$

$$\frac{X_{fft}}{\max(X_{fft})+\epsilon}, \text{ where } \epsilon = 10^{-8} \text{ ensures numerical stability} \quad (48)$$

This transformation effectively captures seismic signal frequency characteristics while preserving scale consistency across earthquake recordings. Model training employed binary cross-entropy loss:

$$\mathcal{L} = -\frac{1}{N}\sum_{i=1}^{N}[y_i \log(\hat{y}_i) + (1 - y_i)\log(1 - \hat{y}_i)] \quad (49)$$

where $y_i$ represents the true label and $\hat{y}_i$ is the predicted probability. This loss function is particularly suitable for our binary classification task of predicting liquefaction occurrence. The model optimization employs the Adam optimizer with a learning rate of $1 \times 10^{-4}$ and weight decay of $1 \times 10^{-3}$ to prevent overfitting. The final classification layer implements a softmax activation function: $p_i = \frac{e^{x_i}}{\sum_j e^{x_j}}$, generating probability distributions over binary outcomes. The model's performance evaluation incorporates multiple metrics, including binary cross-entropy loss and classification accuracy, providing comprehensive assessment of prediction reliability. Implementation utilized PyTorch Lightning framework with GPU acceleration via NVIDIA RTX 4090, ensuring computational efficiency and reproducibility through consistent random seed initialization. Model accuracy was quantified as:

$$\text{Accuracy} = \frac{1}{N}\sum_{i=1}^{N}\mathbb{1}(\text{Argmax}(y_i) = \text{Argmax}(\hat{y}_i)) \quad (50)$$

The model employs a softmax activation function: $p_i = \frac{e^{x_i}}{\sum_j e^{x_j}}$, for generating probability distributions over binary classification outcomes, enabling both decisive predictions and uncertainty quantification. Performance evaluation utilizes multiple metrics, specifically binary cross-entropy loss and classification accuracy, providing comprehensive assessment of the model's predictive capabilities across both positive and negative liquefaction events. The testing protocol implements deterministic evaluation through consistent random seed initialization across data partitioning and training phases. The computational framework leverages PyTorch Lightning with NVIDIA RTX 4090 GPU acceleration, incorporating automatic device detection for optimal resource utilization. This implementation architecture ensures both reproducibility and computational efficiency across diverse computing environments.

    A comprehensive validation framework, integrating k-fold cross-validation and stratified train/test partitioning, was implemented to optimize the proposed neural network architecture. The training methodology addressed data sufficiency and generalization concerns through a dual-phase approach. Initially, a rigorous 10-fold cross-validation protocol systematically partitioned the

dataset into ten segments, employing nine segments for training and one for validation in each iteration cycle. The training protocol utilized 500 epochs with a batch size of 20, optimizing performance metrics for each fold. This cross-validation phase demonstrated robust performance with a mean accuracy of 80% ($\sigma < 7\%$). Performance metric analysis revealed consistent validation results across all folds, with stability indicators remaining within one standard deviation, suggesting robust generalization characteristics.

Subsequently, the dataset underwent a 95%/5% train-validation split for final model optimization, implementing 500 epochs with a batch size of 20, selected to enhance generalization given the constrained dataset dimensions. Model performance was evaluated through multiple complementary metrics: binary cross-entropy loss and classification accuracy. To mitigate overfitting risks, a checkpoint mechanism preserved optimal model states based on validation accuracy peaks, achieving significant performance metrics with an accuracy of 0.9375 and recall of 0.9545 on the validation set. The high accuracy-recall combination demonstrates the model's robust capability in correctly identifying both liquefaction and non-liquefaction cases. This systematic dual-validation approach maintained model stability across heterogeneous geological conditions while demonstrating robust generalization characteristics, as evidenced by consistent performance metrics across validation folds and final testing. The comprehensive evaluation framework and achieved performance metrics support the model's applicability in novel liquefaction susceptibility assessments, particularly within diverse geological and seismic contexts, suggesting reliable generalization to previously unseen cases.

## 3.2 Ablation study

An extensive ablation study was conducted to optimize the model architecture for liquefaction prediction, revealing crucial insights about component contributions and architectural efficiency. The investigation began with baseline component removal tests, where eliminating the ground motion component reduced accuracy from 93.75% to 75.00%, while removing site features led to a 78.79% accuracy, both maintaining the computational cost of 3.305G FLOPs and 1,797,060 parameters. This differential impact suggests that while both components are essential, ground motion data might have a slightly more critical role in prediction accuracy. The study then delved into the attention mechanism configurations, particularly focusing on the soil profile encoder and earthquake signal processor. In the soil profile encoder, experiments with attention heads revealed that while 8 heads achieved 93.75% accuracy (3.306G FLOPs, 1,907,204 parameters), reducing to 1 head dramatically decreased accuracy to 69.70%. This significant performance drop highlights the importance of multiple attention heads in capturing complex soil profile relationships. Similarly, the investigation of attention loops showed that 4 loops could maintain peak accuracy, suggesting a sufficient depth of processing is necessary for optimal performance.

The earthquake signal processor demonstrated even more pronounced sensitivity to attention head configuration. While 8 attention heads maintained 93.75% accuracy, it came at a substantial computational cost of 6.773G FLOPs and 1,851,204 parameters. Reducing to 1 head severely impacted performance, dropping accuracy to 63.64%, despite the improved computational efficiency (1.627G FLOPs, 1,749,380 parameters). This trade-off between performance and computational cost led to a critical optimization challenge. Through systematic evaluation of these various configurations, the study identified an optimal architecture that balances performance and efficiency: 4 attention heads and 2 loops in the soil profile encoder, combined with 2 attention heads in the earthquake signal processor. This optimized configuration achieves the maximum accuracy of 93.75% while maintaining computational efficiency at 3.305G FLOPs and 1,797,060 parameters. This finding is particularly significant as it demonstrates that peak performance can be maintained with a more streamlined architecture, effectively reducing computational complexity without sacrificing prediction accuracy.

The success of this optimized configuration suggests that while multiple attention heads and processing loops are crucial for model performance, there exists a "sweet spot" where additional complexity yields diminishing returns. This balance point effectively captures the necessary feature relationships for accurate liquefaction prediction while maintaining computational efficiency, making the model more practical for real-world applications. The study's findings not only provide valuable insights into the relative importance of different architectural components but also offer practical guidelines for designing efficient deep learning models for geotechnical applications.

Table 1 The results of ablation study

| Model | Accuracy (%) | FLOPs | Total params |
|---|---|---|---|
| Model w/o ground motion | 75.00 | 3.305G | 1,797,060 |
| Model w/o site feature | 78.79 | 3.305G | 1,797,060 |
| 8 head of attention of Soil profile encoders | 93.75 | 3.306G | 1,907,204 |
| 4 loops of attention of soil profile encoder | 93.75 | 3.306G | 1,797,060 |
| 1 head of attention of Soil profile encoders | 69.70 | 3.305G | 1,772,804 |
| 8 head of attention in Earthquake signal processor | 93.75 | 6.773G | 1,851,204 |
| 1 head of attention in Earthquake signal processor | 63.64 | 1.627G | 1,749,380 |
| **Proposed Model** <br> 4 head of attention of Soil profile encoders <br> 2 loops of attention of soil profile encoder <br> 2 head of attention in Earthquake signal processor | 93.75 | 3.305G | 1,797,060 |

## 4. Explainable Artificial intelligent

Model interpretability was examined through SHAP (SHapley Additive exPlanations) values to elucidate both global and local behavioral characteristics of the deep learning model (Fig. 7). The global interpretation provides comprehensive insights into feature importance and their relative contributions to the model's predictive mechanisms, enabling systematic understanding of liquefaction phenomena. Local interpretability analysis generates instance-specific explanations for individual predictions, delineating the precise factors contributing to site-specific liquefaction susceptibility assessments. This dual-level interpretability framework facilitates both macro-scale understanding of model behavior and micro-scale analysis of prediction rationale, enhancing the model's utility for scientific investigation and engineering applications. The SHAP analysis reveals key influential parameters governing liquefaction susceptibility, thereby bridging the gap between complex deep learning architectures and domain-specific geological knowledge.

SHAP, introduced by (Lundberg and Lee, 2017) and extended by (Li et al., 2024), represents a theoretically grounded approach for interpreting predictions from machine learning models through feature attribution. This method quantifies individual feature contributions to model predictions by leveraging concepts from cooperative game theory, specifically Shapley values. In the SHAP framework, features serve as players in a cooperative game where the model's prediction represents the game's value function. The fundamental principle of SHAP lies in its systematic distribution of the prediction value across features based on their marginal contributions. This attribution method has demonstrated particular efficacy in analyzing complex model behaviors across diverse domains. For instance, in linear models, SHAP provides granular decomposition of feature effects, while in non-linear architectures, it captures intricate feature interactions that elude traditional interpretation methods. The computation of SHAP values involves evaluating the marginal contribution of each feature across all possible feature subsets. For a feature ii and prediction function $f(x)$, the SHAP value $\phi_i$ is formally defined as:

$$\phi_i(f) = \sum_{S \subseteq F \setminus \{i\}} \frac{|S|!(|F|-|S|-1)!}{|F|!} x[f(x_{S \cup \{i\}}) - f(x_S)] \qquad (7)$$

where $F$ represents the complete feature set, $S$ denotes a subset of features excluding $i$, and $f_x(S)$ represents the model prediction for feature subset $S$. SHAP values exhibit four fundamental properties that establish their theoretical robustness: efficiency, symmetry (identical contributions yield equivalent SHAP values), dummy (zero-impact features receive null attribution), and additivity (component-wise SHAP values sum to the model-level SHAP value). These properties ensure that SHAP provides a mathematically rigorous framework for model interpretation. The method has gained particular prominence in domains requiring high interpretability standards, such as healthcare diagnostics, financial risk assessment, and regulatory compliance, where model transparency directly impacts stakeholder trust and system accountability.

Global SHAP value analysis revealed that earthquake characteristics (EQ) were the dominant predictor of liquefaction potential, as evidenced by the largest magnitude of SHAP values ranging from -0.4 to +0.4 and the highest position in the feature importance ranking (Fig. 7). This finding is substantiated by the broad distribution of both positive (red) and negative (blue) SHAP values, indicating a complex, non-linear relationship between seismic inputs and liquefaction susceptibility. Following this, spectral parameters at depths of 10, 9, and 6 m (SPT_10, SPT_9, SPT_6) exhibited the next most influential effects on liquefaction occurrence, with SHAP values ranging approximately from -0.2 to +0.2, demonstrating consistent and structured patterns of impact across the dataset.

The shear wave velocity measured at 30 m depth (VS30) emerged as a moderate parameter, as indicated by its mid-range position in the feature hierarchy and SHAP values concentrated within ±0.1. The bidirectional influence of VS30, shown by both positive and negative SHAP values, suggests a complex relationship between wave propagation characteristics and liquefaction susceptibility that warrants further investigation. This observation is particularly noteworthy given the traditional emphasis on VS30 in seismic response analysis. Soil characteristics at various depths (Soil_3, Soil_7, Soil_4) demonstrated moderate influence on liquefaction potential, as evidenced by their intermediate positioning in the SHAP ranking and value ranges typically within ±0.1. The systematic variation in SHAP values across different soil types provides quantitative support for the differential effects of soil composition on liquefaction susceptibility. The data reveals that while soil type remains a significant factor, its influence is secondary to seismic parameters, challenging some traditional assumptions about the primacy of soil characteristics in liquefaction assessment.

Groundwater table depth (WT) and distance to water sources (Dist_Water) showed relatively minor importance, as demonstrated by their lower positioning in the SHAP ranking and more concentrated distribution of impact values. This finding is particularly significant given the historical emphasis on groundwater conditions in liquefaction analysis. The epicentral distance (Dist_epi) and depth parameters similarly demonstrated moderate to low influence, with their SHAP values predominantly clustered around zero, suggesting a more nuanced role in liquefaction triggering than previously postulated.

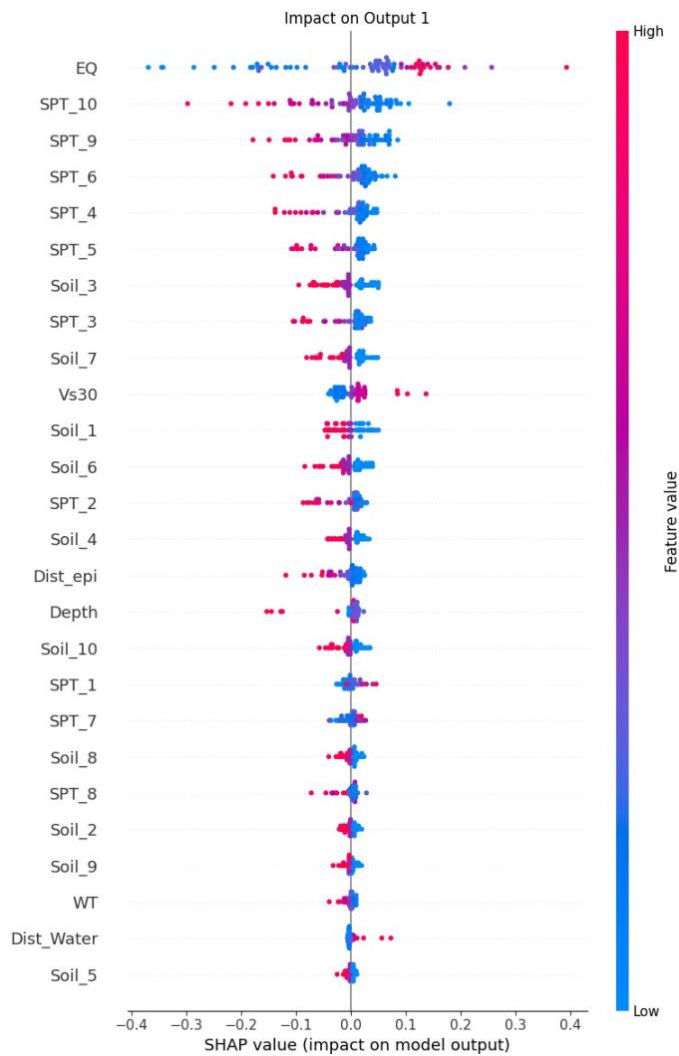

Figure 7 The SHAP value represents the feature effects to liquefaction potential

The absolute SHAP value analysis presented in Figure 8 provides a comprehensive quantitative assessment of feature importance in liquefaction potential prediction. The analysis reveals that earthquake characteristics (EQ) exert the most substantial influence on liquefaction potential, with an absolute SHAP value of 0.1131, significantly exceeding the impact of other parameters. This dominance of seismic input aligns with fundamental principles of soil dynamics and emphasizes the critical role of earthquake loading in triggering liquefaction phenomena. Following the primary seismic influence, spectral parameters at various depths (SPT_10, SPT_9, and SPT_6) demonstrate notable significance, with absolute SHAP values of 0.0570, 0.0397, and 0.0364, respectively. This hierarchical arrangement of spectral parameters suggests a depth-dependent influence on liquefaction susceptibility, with measurements at greater depths (10m) showing more substantial impact than shallower measurements. The systematic decrease in influence with depth provides valuable insights into the spatial distribution of liquefaction sensitivity within soil profiles.

The analysis reveals a moderate influence tier comprising parameters such as SPT_4 through Soil_1, with absolute SHAP values ranging from 0.0320 to 0.0218. Notably, the shear wave velocity (VS30) falls within this range with a value of 0.0228, indicating its moderate yet significant role in liquefaction assessment. This finding suggests that while wave propagation characteristics contribute to liquefaction susceptibility, their influence is secondary to direct seismic loading parameters. Particularly noteworthy is the relatively minor influence of traditionally emphasized parameters such as water table depth (WT) and distance to water sources (Dist_Water), which exhibit low absolute SHAP values of 0.0055 and 0.0052, respectively. This quantitative evidence challenges conventional assumptions about the primacy of groundwater conditions in liquefaction assessment and suggests a need to reevaluate traditional weighting of these parameters in practical applications. The gradual degradation in absolute SHAP values across the feature spectrum, spanning more than an order of magnitude from the highest (0.1131) to lowest (0.0050) values, indicates a continuous rather than discrete hierarchy of importance. This smooth transition suggests that comprehensive liquefaction assessment should consider multiple parameters while appropriately weighting their relative contributions based on their quantified importance. These findings have significant implications for both theoretical understanding and practical applications in geotechnical engineering. The clear quantitative ranking provided by absolute SHAP values supports a shift toward more seismic-centric approaches to liquefaction assessment, while maintaining consideration of soil and groundwater parameters in proportion to their demonstrated influence. This evidence-based hierarchy of importance can inform the development of more refined and accurate liquefaction prediction methodologies.

The interaction patterns revealed by the SHAP analysis, particularly the varying distributions and color gradients across different parameter combinations, provide strong evidence for the complexity of liquefaction mechanisms. The dominance of seismic parameters over geotechnical factors is consistently supported by the magnitude and distribution of SHAP values across the feature set. This comprehensive quantitative assessment suggests that traditional univariate approaches to liquefaction evaluation may need revision to better reflect the multivariate nature of the phenomenon. These findings are further reinforced by the systematic patterns observed in the SHAP value distributions, which indicate structured relationships between parameters rather than random associations. This analysis contributes to the field by providing quantitative evidence for the relative importance of different parameters in liquefaction assessment, supported by systematic SHAP value patterns and distributions. The findings suggest a need for refined approaches to liquefaction susceptibility evaluation that better account for the dominant role of seismic parameters while appropriately weighing the contributions of soil and groundwater conditions.

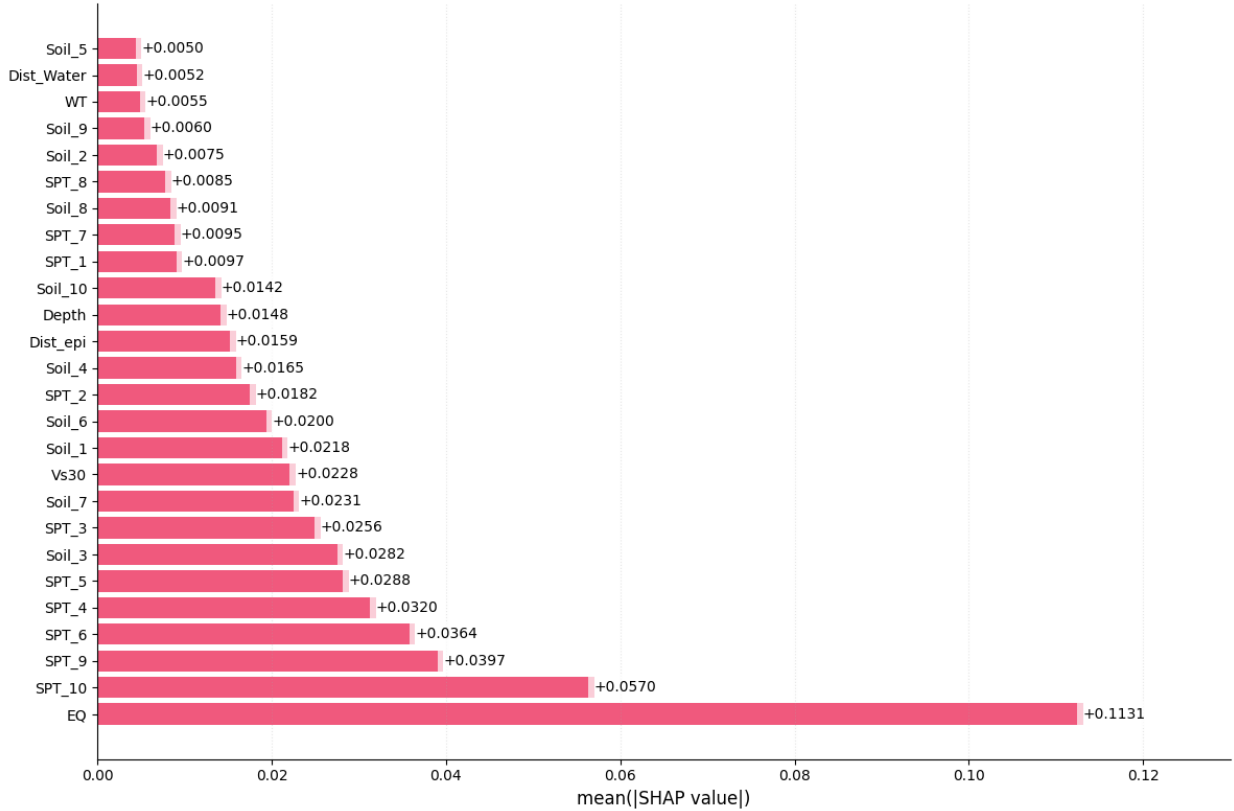

Figure 8 The Absolute SHAP value represents the feature effects to liquefaction potential

Beyond its global interpretability capabilities, SHAP offers fine-grained attribution analysis at the instance level, enabling precise interpretation of site-specific liquefaction predictions. This local interpretability framework decomposes individual predictions into feature-wise contributions, quantifying the relative influence of each geotechnical and seismic parameter on location-specific liquefaction susceptibility. Such granular analysis not only illuminates the mechanistic drivers of liquefaction risk but also informs targeted mitigation strategies through systematic feature importance decomposition. To validate this local interpretation approach, we strategically selected test locations representing diverse geotechnical conditions from our validation dataset, demonstrating the practical utility of SHAP-based feature attribution in real-world geotechnical engineering applications.

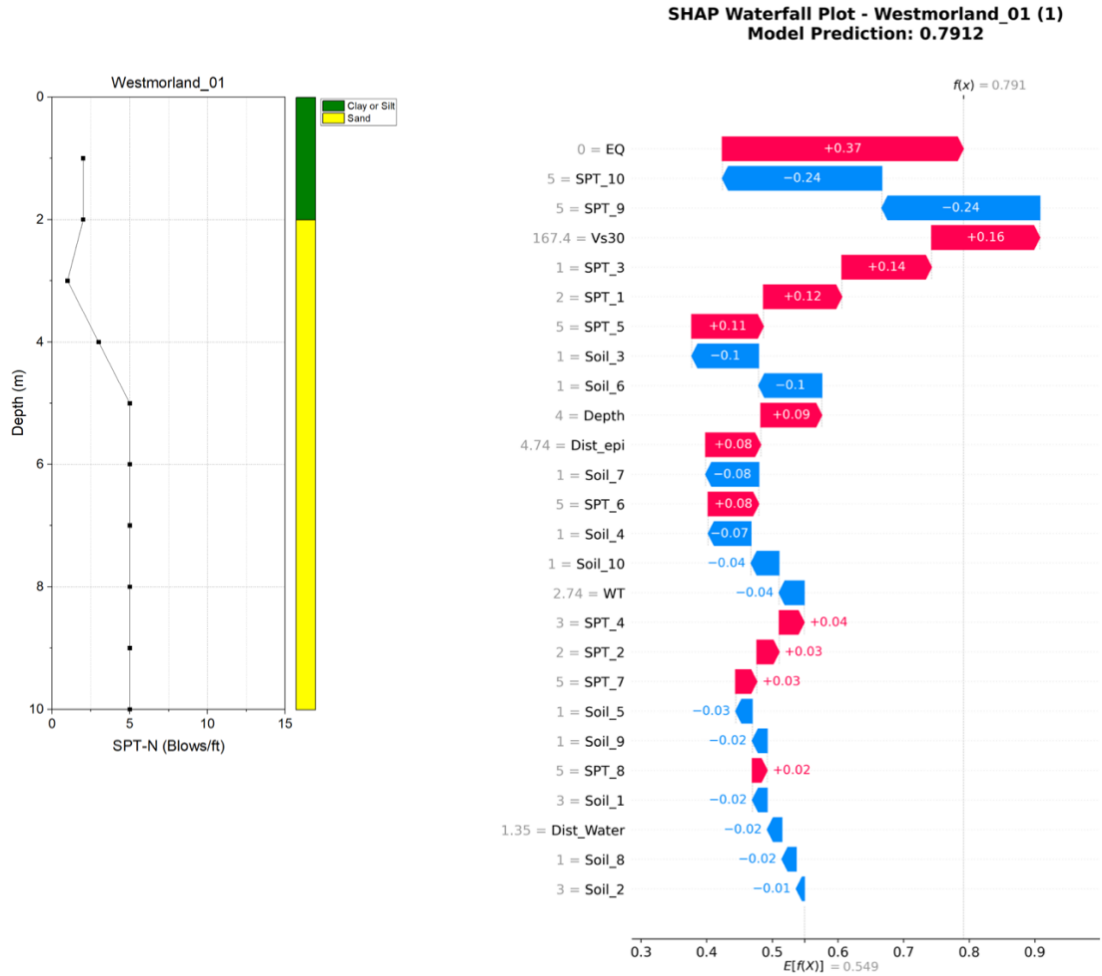

Figure 9 The local SHAP value of the test data Westmorland_01

Local SHAP analysis was performed on the Westmorland_01 test case (Fig. 9). The geotechnical investigation revealed a stratified soil profile comprising clay/silt in the upper 2 meters underlain by sand deposits extending to 10 m depth, as evidenced by the borehole log. Standard Penetration Test (SPT-N) measurements exhibited a non-linear depth-dependent distribution, with values ranging from approximately 2 blows/ft near the surface to 15 blows/ft at terminal depth. The SHAP-based liquefaction probability assessment yielded f(x) = 0.791, substantially exceeding the E[f(x)] baseline of 0.549, indicating elevated liquefaction susceptibility. Analysis of SPT measurements revealed complex non-linear relationships with liquefaction susceptibility. At depths of 9 and 10 meters, SPT_9 and SPT_10 yielded significant negative SHAP values (-0.24 each), while measurements at intermediate depths (SPT_3, SPT_1, SPT_5) showed positive SHAP contributions (+0.14, +0.12, +0.11 respectively). These contrasting SHAP values indicate a non-linear relationship between SPT measurements and liquefaction potential across different depths, suggesting that identical SPT values may have different implications for liquefaction susceptibility depending on their spatial context. The shear wave velocity parameter (Vs30 = 167.4 m/s) demonstrated a notable positive SHAP value (+0.16), though this relationship should also be interpreted within the context of potential non-linear interactions with other soil parameters rather than as a direct correlation.

Spatial parameters demonstrated varying influences: epicentral distance (Dist_epi = 4.74) showed a moderate positive contribution (SHAP value: +0.08), while water table depth (WT = 2.74) exhibited minimal impact (SHAP value: -0.04). Soil classification parameters (Soil_1 through Soil_10) generally showed modest SHAP values ranging from -0.10 to +0.02, indicating their secondary role in liquefaction prediction for this case. These quantitative relationships provide insights for site-specific liquefaction hazard assessment, particularly highlighting the complex depth-dependent behavior of SPT resistance and its correlation with liquefaction susceptibility. The findings suggest that simplified depth-averaged approaches may not adequately capture the stratified nature of liquefaction potential, emphasizing the importance of depth-resolved analysis in geotechnical engineering practice.

## 5. Application to real case study

Model validation was performed using seismic waveform data from the January 1, 2024 Noto Peninsula earthquake (Mw 7.5, MJMA 7.6) in Japan. The mainshock epicenter was located at 37.5°N, 137.2°E, approximately 6 km NNE of Suzu City, Ishikawa Prefecture. The event generated severe ground motions that registered maximum intensities of Shindo 7 on the Japan Meteorological Agency (JMA) scale and X-XI on the Modified Mercalli Intensity (MMI) scale. Significant ground deformation was observed across multiple municipalities in the northern Noto region, including Suzu, Wajima, Noto Town, and Anamizu, with measurable effects extending to adjacent Toyama and Niigata prefectures. The severity of ground deformation is evidenced by widespread soil liquefaction effects, as shown in Figure 10, where sewage manhole covers were significantly uplifted due to increased pore water pressure in the liquefied soil layers. These documented cases of infrastructure damage, particularly in urban areas with paved surfaces and underground utilities, provide crucial validation data for understanding the spatial distribution and intensity of liquefaction phenomena. As this seismic event occurred after our model development phase, these data provide an independent test set for evaluating model performance under extreme ground motion conditions not represented in the training dataset. This earthquake's exceptional magnitude and complexity present a rigorous challenge for validating model robustness.

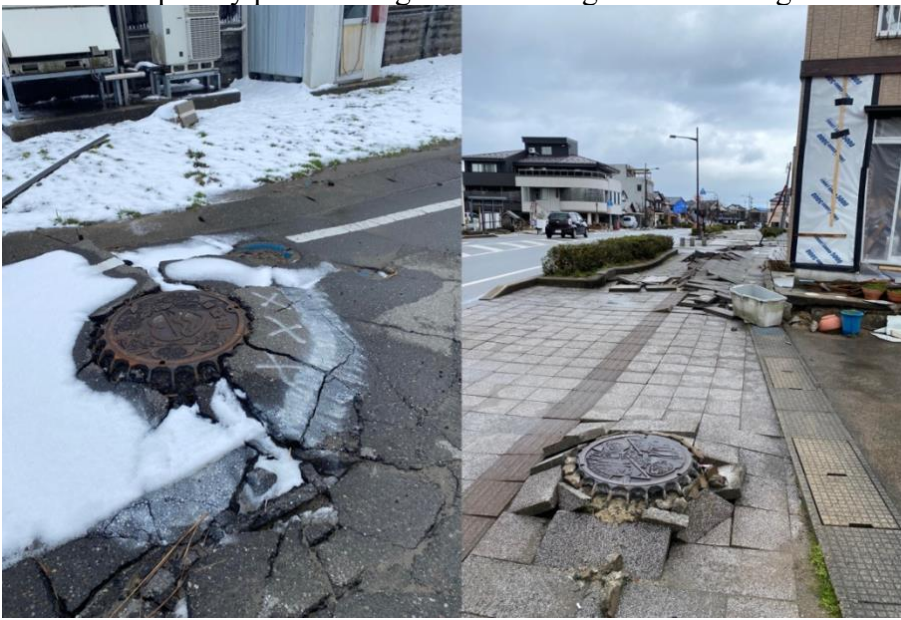

Figure 10 Soil liquefaction damage from the Noto earthquake. The sewage manhole covers were uplifted due to the liquefaction effect.

Field investigations at site ISK002 (37.4413°N, 137.2908°E) in Noto documented extensive liquefaction-induced ground deformation, which was subsequently analyzed through conventional methods and machine learning algorithms (National Research Institute for Earth Science and Disaster Resilience, 2019). Quantitative analysis using SHAP (SHapley Additive exPlanations) methodology yielded a liquefaction probability of 0.785 (Fig. 11). Stratigraphic analysis revealed a heterogeneous subsurface profile comprising an uppermost clay/silt unit, underlain by a primary sand stratum extending to approximately 9 m depth, followed by silty sand deposits. Standard Penetration Test (SPT) measurements indicated vertical variation in soil density, with N-values ranging from approximately 0 blows/ft at the surface to 50 blows/ft at depth. SHAP feature importance analysis demonstrated that seismic parameters exhibited the highest positive correlation (+0.04) with liquefaction susceptibility. Stratigraphic units 7, 3, and 1 each contributed +0.02 to the probability calculation, while SPT measurements (layers 4-9) consistently demonstrated +0.01 positive correlations. These quantitative assessments corresponded with observed surface manifestations, including sand boils and artesian groundwater discharge. Epicentral distance and groundwater depth parameters exhibited minimal negative correlations, indicating the model's capacity to integrate multiple variables with greater precision than conventional analytical methods. The machine learning algorithm effectively characterized the complex seismic-soil interactions, resulting in a final probability value (0.785) that exceeded the baseline prediction (E[f(X)] = 0.549). This enhanced prediction demonstrated superior correlation with documented damage patterns, particularly in proximity to hydraulic infrastructure. These results suggest that machine learning methodologies may provide more accurate assessments of liquefaction potential in complex geological environments where traditional analytical approaches may underestimate risk factors.

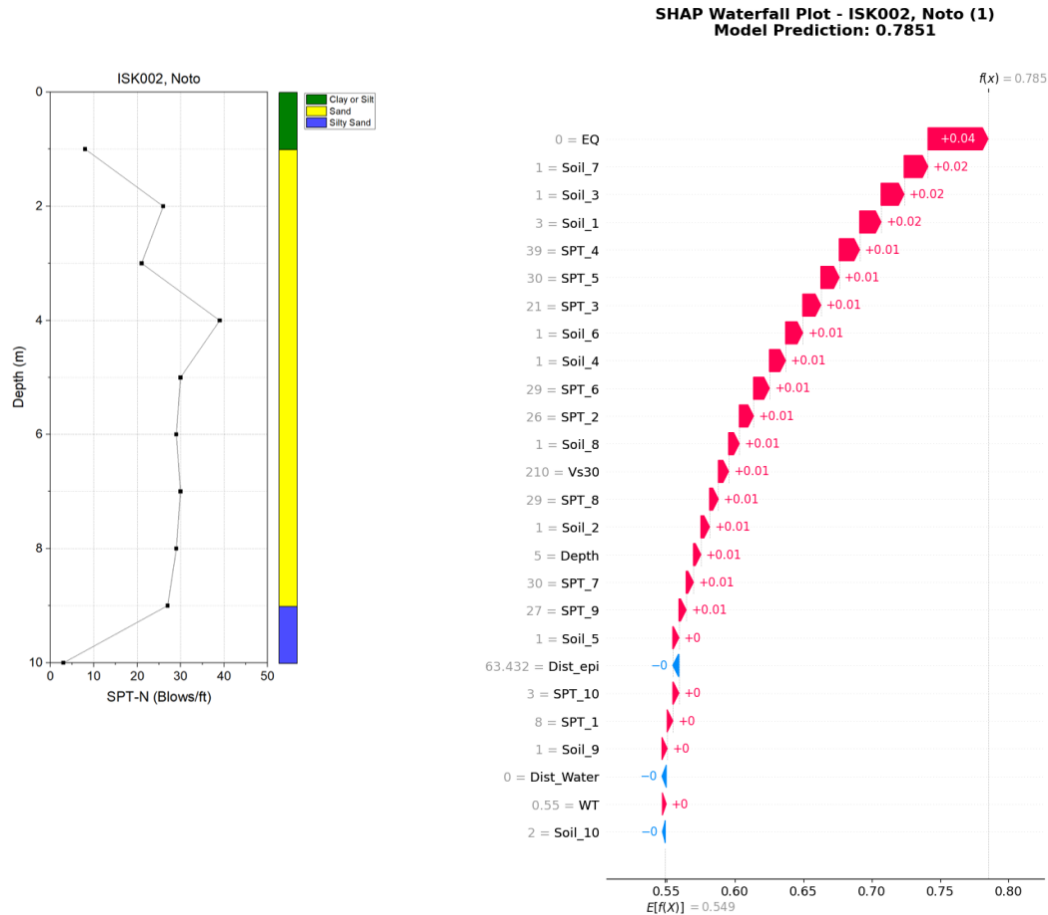

Figure 11 The soil profile and SHAP waterfall plot for ISK002 site during the 2024 Noto Earthquake, showing SPT-N values and individual parameter contributions to liquefaction prediction.

The conventional analysis of potential of liquefaction in each layer was present in Fig. 12. (Seed and Idriss, 1971). It is a fundamental approach for assessing soil liquefaction potential during earthquakes. This simplified procedure evaluates liquefaction susceptibility by comparing the seismic demand placed on a soil layer with its capacity to resist liquefaction. The seismic demand is quantified through the Cyclic Stress Ratio (CSR), which accounts for the ground motion intensity, depth effects, and stress conditions in the soil. The CSR calculation incorporates the peak ground acceleration, total and effective vertical stresses, and a depth-dependent stress reduction coefficient. The soil's resistance to liquefaction is expressed as the Cyclic Resistance Ratio (CRR), which is typically determined from Standard Penetration Test (SPT) N-values. The potential for liquefaction is assessed by calculating the Factor of Safety (FS), which is the ratio of CRR to CSR. A factor of safety greater than 1.0 indicates that the soil layer is unlikely to liquefy, while values less than 1.0 suggest liquefaction is likely to occur. The analysis results are typically presented in graphical form, showing the variation of CSR, CRR, and factor of safety with depth, allowing engineers to identify potentially liquefiable zones within the soil profile. While the Seed and Idriss method has become a standard practice in geotechnical engineering, it has certain limitations. The method was developed primarily based on observations from California earthquakes and is most applicable to clean sands. It requires careful judgment in selecting input parameters and represents a simplified approach that may not capture all aspects of complex soil behavior during seismic loading. Despite these limitations, the method remains widely used due to its practical nature and extensive validation through field observations.

The liquefaction assessment utilizing the conventional method presents a comprehensive analysis of soil stability through the examination of (1-FOS) values across varying depths up to 10 meters (Fig. 11). The graphical representation demonstrates distinctive patterns of liquefaction susceptibility throughout the soil profile. The surface layer at 1 m depth exhibits a significant (1-FOS) value of 0.7488, indicating substantial liquefaction potential. This high value suggests that the near-surface material may be particularly vulnerable to seismic-induced deformation. The intermediate depths, ranging from 4 m to 9 m, display relatively consistent (1-FOS) values approximating 0.48-0.52, suggesting moderate liquefaction susceptibility. This uniformity in values indicates a relatively homogeneous response to seismic loading within these intermediate layers. However, the most critical condition is observed at the 10m depth, where the (1-FOS) value reaches 0.9331, representing the highest liquefaction potential within the analyzed profile. This pronounced susceptibility at depth warrants particular attention in geotechnical design considerations. The interpretation framework for (1-FOS) values establishes that measurements approaching 1.0 indicate enhanced liquefaction potential, while values closer to 0 suggest greater stability against liquefaction. The analysis reveals that values exceeding 0.5 generally indicate concerning levels of liquefaction susceptibility, which is evident in several layers throughout the profile. Notably, the 2m depth demonstrates the lowest value of 0.4256, potentially representing the most stable zone within the analyzed soil column. This comprehensive assessment provides crucial insights for geotechnical design and hazard mitigation strategies, particularly in seismically active regions where liquefaction poses significant risks to infrastructure stability.

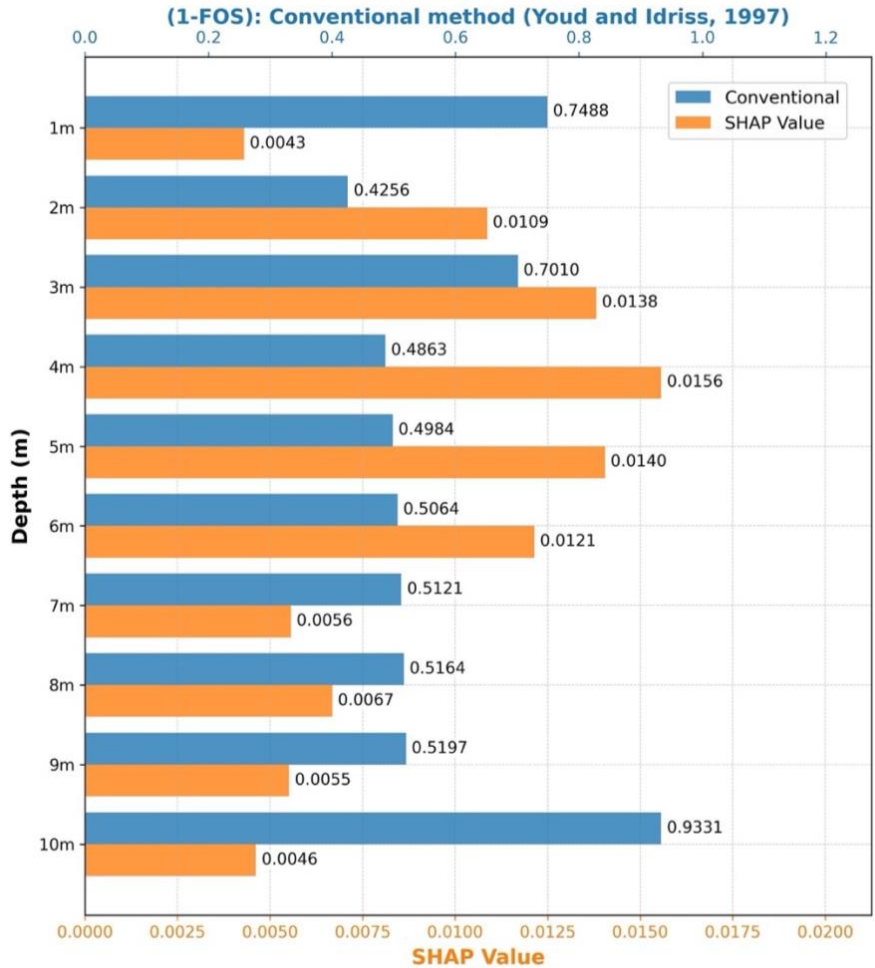

Figure 12 SHAP Values from proposed model and Liquefaction Probability (1 - Factor of Safety) from conventional method (Youd and Idriss, 1997) at Different Soil Depths: Case Study of the Noto Earthquake

SHAP value analysis reveals that the earthquake parameter (EQ) is the dominant contributor to liquefaction susceptibility, increasing the model probability by 0.04 from the base value of 0.549 (Fig. 12). Soil characteristics at three depths (Soil_7, Soil_3, and Soil_1) emerge as the next most influential parameters, each contributing +0.02 to the prediction. The Standard Penetration Test (SPT) measurements at middle depths, particularly SPT_4 (39 blows/ft), SPT_5 (30 blows/ft), and SPT_3 (21 blows/ft), demonstrate higher influence than other SPT readings, each adding +0.01 to the prediction probability. Parameters showing minimal impact (near zero or slight negative contribution) include distance to epicenter (Dist_epi), distance to water (Dist_Water), water table depth (WT), and deep soil classification (Soil_10), with the model achieving a final prediction value of 0.785. Notably, the transformer model's interpretation of SPT values presents an interesting departure from conventional liquefaction assessment approaches. While traditional methods typically correlate liquefaction susceptibility with low SPT values in isolation, the transformer model processes the entire soil profile as a sequence, revealing the significance of layer contrasts and transitions. High SPT values occurring within the complex profile of alternating clay and silty sand layers suggest that stiff layers sandwiched between softer materials create impedance contrasts that can trap seismic energy and lead to strain localization.

This holistic analysis indicates that liquefaction susceptibility is more nuanced than previously considered, with the arrangement and interaction of soil layers throughout the profile playing a crucial role in seismic behavior. These findings suggest that future liquefaction assessments should consider not only individual soil parameters but also their stratigraphic context and layer interactions within the complete soil profile.

Figure 12 reveals significant differences in both the scale and distribution of values between the two methods. The conventional method (1-FOS) shows relatively high values ranging from 0.4256 to 0.9331, with particularly high liquefaction potential at 1m (0.7488), 3m (0.7010), and 10m (0.9331) depths, indicating these layers are more susceptible to liquefaction according to traditional calculations. However, these high variations between adjacent layers may not accurately represent the continuous nature of soil behavior. In contrast, the proposed model's SHAP values, while operating on a much smaller scale (0.0043 to 0.0156), demonstrate a more gradual pattern of change. It's important to note that these SHAP values do not represent overall liquefaction probability, but rather quantify each layer's contribution to the liquefaction potential in the soil profile. The highest contributions are concentrated in the middle layers, particularly at 4m (0.0156) and 3m (0.0138), suggesting these depths have the strongest influence on the overall liquefaction behavior. The contribution values then systematically decrease towards deeper layers, with values dropping to 0.0046 at 10m depth, indicating a reduced influence on the liquefaction potential. This progressive reduction in contribution values with depth aligns better with the physical understanding of how liquefaction typically develops in soil profiles, where upper layers generally play a more critical role in initiating the process. The proposed method demonstrates several advantages over the conventional approach: it considers the interconnected nature of soil layers rather than treating them independently, accounts for the sequential nature of soil deposits which better reflects geological formation processes, reduces the artificial variations that can arise from independent layer calculations, and provides a more nuanced understanding of how each layer contributes to the overall liquefaction potential. Additionally, by incorporating machine learning techniques, the proposed method can capture complex non-linear relationships between soil parameters that may be overlooked in conventional deterministic calculations.

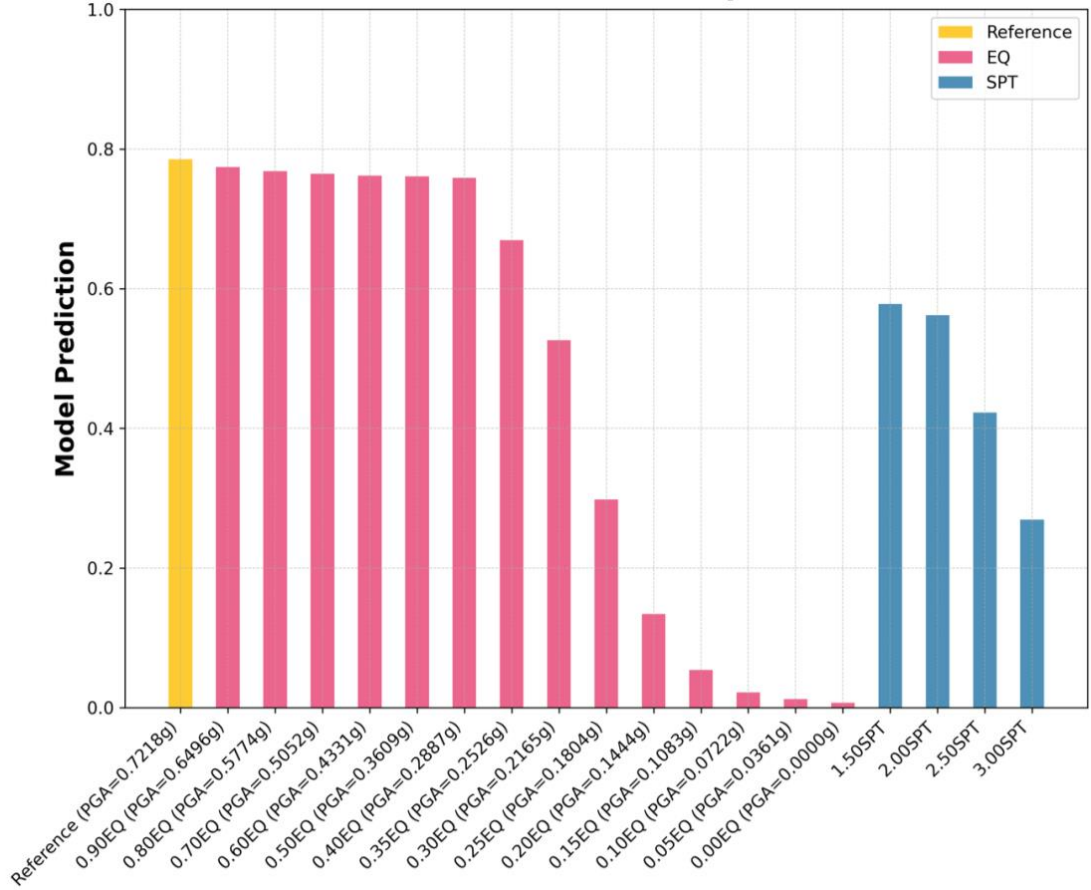

Figure 13 Sensitivity analysis comparing model predictions of liquefaction probability across varying PGA (Peak Ground Acceleration) values and SPT (Standard Penetration Test) measurements.

    The sensitivity analysis performed on the liquefaction prediction model demonstrates its robustness and practical applicability through systematic variation of two critical parameters: ground motion intensity and soil resistance characteristics (Fig. 13). The reference case, with a peak ground acceleration (PGA) of 0.7218g, served as the baseline for the earthquake motion analysis. A systematic reduction in PGA was implemented through multiplication factors, creating a spectrum of scenarios ranging from full intensity (1.0) to complete attenuation (0.0). The results reveal distinct threshold behaviors in the liquefaction response. At high PGA levels (0.5-0.7g), the model predicts liquefaction probabilities consistently above 0.75, indicating high susceptibility. A notable transition occurs as the PGA decreases, with the first critical threshold appearing at approximately 0.2887g, where the probability drops to 0.67. This is followed by a more significant threshold at 0.2165g, where the probability falls below 0.5, conventionally considered the boundary between likely and unlikely liquefaction occurrence. The probability continues to decrease exponentially with further PGA reduction, approaching zero at very low acceleration levels.

    The second aspect of the sensitivity analysis focused on the Standard Penetration Test (SPT) values, representing soil resistance to liquefaction (Fig. 13). The analysis employed multiplication factors applied to the reference SPT profile, effectively simulating various degrees

of soil improvement. The results demonstrate a clear inverse relationship between SPT values and liquefaction probability. When the SPT values are increased to twice the reference values (2.0×SPT), the liquefaction probability decreases below 0.5, indicating effective mitigation of liquefaction risk. This finding has direct implications for ground improvement specifications, providing quantitative targets for soil densification programs. The model's capability to evaluate both uniform and depth-specific variations in soil properties makes it particularly valuable for engineering applications. This flexibility allows practitioners to optimize ground improvement strategies by targeting specific layers or depths where improvement would be most effective. For instance, engineers can evaluate scenarios where only critical layers are treated, potentially leading to more cost-effective ground improvement solutions while maintaining adequate safety margins against liquefaction.

The sensitivity analysis also highlights the model's utility in risk assessment and mitigation planning. By quantifying the relationship between PGA, soil resistance, and liquefaction probability, engineers can develop performance-based criteria for ground improvement projects. Furthermore, the clear identification of threshold values for both seismic loading (PGA) and soil resistance (SPT) provides rational bases for developing site-specific improvement criteria and evaluating the effectiveness of different ground improvement alternatives. This comprehensive understanding of parameter sensitivity enables more informed decision-making in geotechnical earthquake engineering practice, particularly in regions where liquefaction hazard is a significant concern. The comparative assessment suggests that integrating both approaches could enhance liquefaction hazard evaluation by combining SHAP's parameter importance quantification with CSR's established stress-based framework. This integration particularly benefits sites with complex stratigraphy, where traditional methods alone may not fully capture the interaction between seismic loading and soil response characteristics. The complementary nature of these methods underscores the potential for machine learning interpretability tools to augment, rather than replace, conventional geotechnical analyses in liquefaction assessment practices.

## 5.1 Web base application

We developed the web-based Liquefaction Probability Calculator representing a significant advancement in geotechnical engineering analysis, combining machine learning capabilities with practical engineering applications (Fig. 14). The application features an intuitive user interface that allows engineers to upload Excel files containing site and seismic data. Upon file upload, the system rapidly processes the data to generate liquefaction probability predictions for multiple samples simultaneously, providing results on a scale from 0 to 1, where higher values indicate greater liquefaction susceptibility. A particularly innovative aspect of the calculator is its integration of SHAP, which provides transparent insights into the model's decision-making process. As demonstrated in the sample analysis shown in Fig. 14, the system evaluates key parameters such as earthquake characteristics (EQ), soil depth, and Standard Penetration Test values. The SHAP visualization effectively communicates how each factor contributes to the final prediction through a color-coded system, where red indicates positive contributions and blue represents negative influences on liquefaction probability.

The implementation of this tool on the Hugging Face platform [https://huggingface.co/spaces/Sompote/Liquefaction_prediction] ensures widespread accessibility while maintaining robust performance. The application's architecture, built using the Streamlit library, enables seamless processing of complex geotechnical datasets. This technological framework supports rapid analysis, typically completing calculations within seconds, regardless of the dataset's complexity. For practicing engineers, this means the ability to efficiently evaluate multiple site conditions and earthquake scenarios, leading to more informed decision-making in ground improvement and site development projects. The calculator's practical significance extends beyond mere probability calculations. By providing detailed insights into the factors contributing to liquefaction risk, engineers can develop more targeted and cost-effective ground improvement strategies. The tool's ability to handle multiple site conditions and earthquake scenarios simultaneously makes it particularly valuable for large-scale projects or regional assessments. This represents a substantial improvement over traditional methods, offering a more nuanced and comprehensive approach to liquefaction risk assessment while maintaining accessibility for common engineering applications.

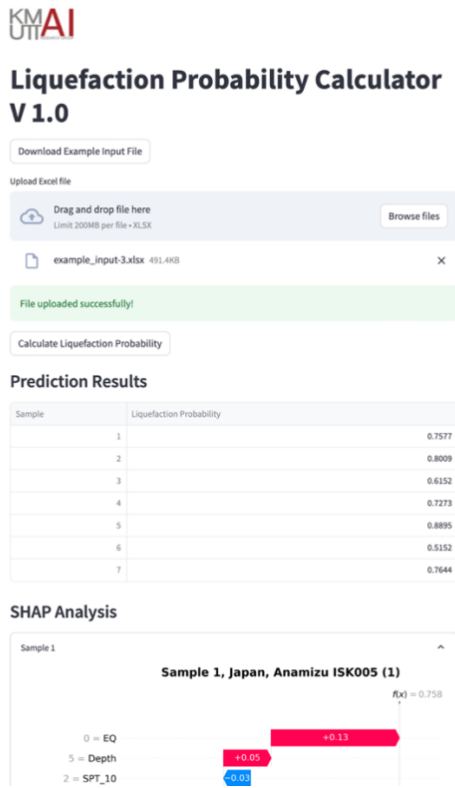

Fig. 14 The web-based application for evaluation of liquefaction [https://huggingface.co/spaces/Sompote/Liquefaction_prediction]

## 6 Discussion

The findings of this research demonstrate significant potential for advancing liquefaction prediction methodologies while highlighting important areas for future development. The transformer-based architecture's achievement of 93.75% accuracy represents a substantial improvement over traditional methods, particularly in its ability to integrate multiple data streams and provide interpretable results. The successful validation against the 2024 Noto Peninsula earthquake underscores the model's practical utility in real-world applications. However, several aspects warrant further investigation to enhance the model's robustness and broader applicability. The current implementation could be strengthened through expansion of the training dataset beyond the existing 165 case histories. While this dataset has proven sufficient for demonstrating the model's capabilities, a larger and more geographically diverse collection of case histories would enhance the model's generalizability across varied geological settings. The integration of additional data from ongoing seismic monitoring networks and post-earthquake investigations would be particularly valuable in this regard. Furthermore, the establishment of standardized protocols for data collection and liquefaction classification would significantly improve the consistency of model training and validation.

The implementation of artificial intelligence for liquefaction prediction has evolved significantly, with our model demonstrating notable advancements over recent studies (Guo et al., 2025; Hsiao et al., 2024; Long et al., 2025). Our model's superior performance, achieving 93.75 % accuracy across diverse geological conditions and seismic intensities, represents a significant improvement in prediction reliability. This enhancement can be attributed to several key methodological innovations in our approach. A critical limitation of previous methodologies was their reliance on simplified, single-dimensional tensor inputs, which failed to capture the complex, sequential nature of soil profiles and seismic events. Many earlier studies either used averaged soil parameters or excluded comprehensive soil profile data altogether, resulting in models with limited applicability across different geological contexts. This oversimplification of input data has been a significant barrier to developing globally applicable liquefaction prediction models. Our methodology addresses these limitations through two principal innovations. First, we implemented transformer-based encoding for soil profile data, enabling the model to capture the intricate layering and spatial relationships within soil structures. Second, our incorporation of earthquake data in the frequency domain allows for a more comprehensive representation of seismic characteristics. These enhancements enable our model to capture complex patterns and relationships that were previously overlooked, contributing to its improved generalization capabilities across diverse geological settings.

The feature importance analysis conducted using SHAP technique revealed patterns consistent with established geotechnical understanding and previous research findings research (Hsiao et al., 2024). Earthquake ground motion emerged as the dominant predictor of liquefaction potential, which aligns with fundamental seismic principles. Secondary factors, including distance to water sources and groundwater table depth, showed measurable but comparatively smaller influences on liquefaction predictions. This hierarchy of influential factors not only validates our model's learning process but also provides valuable insights for geotechnical engineering applications. These improvements in model architecture and feature representation demonstrate the potential for advanced machine learning techniques to enhance our understanding and prediction of soil liquefaction phenomena. The ability to accurately predict liquefaction across diverse geological conditions represents a significant advancement in geotechnical earthquake engineering, with practical implications for seismic hazard assessment and infrastructure planning.

Several promising directions for future research emerge from this work. First, the incorporation of real-time seismic data streams could enable dynamic updating of liquefaction probability assessments during earthquake events, enhancing the model's utility for emergency response applications. Second, the extension of the transformer architecture to predict not only liquefaction occurrence but also its spatial extent and severity would provide more comprehensive hazard assessments. Third, the development of transfer learning approaches could help address the challenge of limited data availability in specific regions while leveraging knowledge from better-documented areas.

Additional research opportunities include the integration of satellite-based ground deformation measurements, enhancement of the frequency-domain analysis to better capture nonlinear soil behavior, and development of uncertainty quantification methods for model predictions. The incorporation of ground motion uncertainty through probabilistic seismic hazard analysis would also strengthen the model's applicability in forward-looking risk assessments. Finally, investigation of more sophisticated interpretability techniques could provide deeper insights into the physical mechanisms underlying the model's predictions, potentially leading to new understanding of liquefaction phenomena.

This research establishes a foundation for next-generation liquefaction hazard assessment tools while identifying clear pathways for continued advancement. The challenges identified throughout this study do not diminish its significant contributions but rather highlight the rich opportunities for further research in this critical area of geotechnical engineering. Continued development along these lines promises to enhance our ability to predict and mitigate liquefaction hazards worldwide.

## 7. Conclusion

This research advances the state-of-the-art in earthquake-induced liquefaction prediction through the development of an interpretable multi-modal transformer architecture. Our model achieves 93.75% prediction accuracy across diverse geological settings. The architecture's success stems from its innovative parallel processing of three distinct data streams: frequency-domain seismic characteristics processed through Fast Fourier Transform, depth-wise soil profiles encoded using principles from natural language processing, and site-specific features including groundwater conditions and shear wave velocity measurements. This integration enables comprehensive analysis of liquefaction susceptibility through specialized encoding mechanisms, advancing beyond conventional discrete analysis approaches.

Central to our contribution is the implementation of a novel interpretability framework using SHAP (SHapley Additive exPlanations) values. This analysis revealed seismic parameters as the dominant predictor, with soil profiles at depths of 5-6m showing the next highest influence, while Vs30 measurements and groundwater conditions demonstrated moderate influence. The quantitative attribution of feature importance bridges a crucial gap between advanced AI architectures and practical engineering applications, addressing a long-standing challenge in geotechnical machine learning implementations.

The model's practical utility was demonstrated through comprehensive validation using the January 2024 Noto Peninsula earthquake data. Our evaluation included direct validation and sensitivity analysis, revealing superior performance, particularly in cases where traditional methods underestimate liquefaction severity. The sensitivity studies showed robust model responses to variations in ground motion intensity and soil resistance parameters, providing valuable insights for practical applications. Furthermore, the successful implementation of a web-based platform enables automated processing of multiple sites simultaneously, facilitating rapid regional assessments during seismic events.

Looking forward, this research opens several promising avenues for future development. Integration of real-time seismic monitoring data and satellite-based deformation measurements could enhance the model's predictive capabilities, while expansion to three-dimensional spatial analysis would provide more comprehensive hazard assessments. The framework established here provides a foundation for next-generation liquefaction hazard assessment tools, potentially transforming how we approach seismic risk evaluation and mitigation worldwide. By combining sophisticated AI techniques with robust explainability mechanisms, our methodology establishes a new paradigm in geotechnical engineering that balances advanced predictive capabilities with practical interpretability requirements, essential for the broader adoption of AI-driven approaches in critical infrastructure applications

## 8. Acknowledgments

The authors gratefully acknowledge the financial support provided by King Mongkut's University of Technology Thonburi (KMUTT), Thailand Science Research and Innovation (TSRI), and National Science, Research and Innovation Fund (NSRF) Fiscal year 2026 under the project titled "Application of artificial intelligence and advanced computation for infrastructure projects".

## 9. References

Abbaszadeh Shahri, A., 2016. Assessment and Prediction of Liquefaction Potential Using Different Artificial Neural Network Models: A Case Study. Geotech Geol Eng 34, 807–815. https://doi.org/10.1007/s10706-016-0004-z

Abdollahi, A., Li, D., Deng, J., Amini, A., 2024. An explainable artificial-intelligence-aided safety factor prediction of road embankments. Engineering Applications of Artificial Intelligence 136, 108854. https://doi.org/10.1016/j.engappai.2024.108854

Andrus, R.D., Stokoe Ii, K.H., 2000. Liquefaction Resistance of Soils from Shear-Wave Velocity. J. Geotech. Geoenviron. Eng. 126, 1015–1025. https://doi.org/10.1061/(ASCE)1090-0241(2000)126:11(1015)

Bol, E., 2023. A new approach to the correlation of SPT-CPT depending on the soil behavior type index. Engineering Geology 314, 106996. https://doi.org/10.1016/j.enggeo.2023.106996

Boulanger, R.W., Idriss, I.M., 2016. CPT-Based Liquefaction Triggering Procedure. J. Geotech. Geoenviron. Eng. 142, 04015065. https://doi.org/10.1061/(ASCE)GT.1943-5606.0001388

Brandenberg, S.J., Zimmaro, P., Stewart, J.P., Kwak, D.Y., Franke, K.W., Moss, R.E., Çetin, K.Ö., Can, G., Ilgac, M., Stamatakos, J., Weaver, T., Kramer, S.L., 2020. Next-generation liquefaction database. Earthquake Spectra 36, 939–959. https://doi.org/10.1177/8755293020902477

Cetin, K.O., Seed, R.B., Kayen, R.E., Moss, R.E.S., Bilge, H.T., Ilgac, M., Chowdhury, K., 2018. SPT-based probabilistic and deterministic assessment of seismic soil liquefaction triggering hazard. Soil Dynamics and Earthquake Engineering 115, 698–709. https://doi.org/10.1016/j.soildyn.2018.09.012

Chen, X., Ma, C., Ren, Y., Lei, Y.-T., Huynh, N.Q.A., Narayan, S.W., 2023. Explainable artificial intelligence in finance: A bibliometric review. Finance Research Letters.

Demir, S., Sahin, E.K., 2022. Comparison of tree-based machine learning algorithms for predicting liquefaction potential using canonical correlation forest, rotation forest, and random forest based on CPT data. Soil Dynamics and Earthquake Engineering 154, 107130. https://doi.org/10.1016/j.soildyn.2021.107130

Fahim, A.K.F., Rahman, Md.Z., Hossain, Md.S., Kamal, A.S.M.M., 2022. Liquefaction resistance evaluation of soils using artificial neural network for Dhaka City, Bangladesh. Nat Hazards 113, 933–963. https://doi.org/10.1007/s11069-022-05331-w

Goh, A.T.C., 1996. Neural-Network Modeling of CPT Seismic Liquefaction Data. J. Geotech. Engrg. 122, 70–73. https://doi.org/10.1061/(ASCE)0733-9410(1996)122:1(70)

Goh, A.T.C., Goh, S.H., 2007. Support vector machines: Their use in geotechnical engineering as illustrated using seismic liquefaction data. Computers and Geotechnics 34, 410–421. https://doi.org/10.1016/j.compgeo.2007.06.001

Guo, H., Zhang, C., Fang, H., Rabczuk, T., Zhuang, X., 2025. Deep learning to evaluate seismic-induced soil liquefaction and modified transfer learning between various data sources. Underground Space S246796742400134X. https://doi.org/10.1016/j.undsp.2024.08.010

Hanna, A.M., Ural, D., Saygili, G., 2007. Evaluation of liquefaction potential of soil deposits using artificial neural networks. Engineering Computations 24, 5–16. https://doi.org/10.1108/02644400710718547

Hsein Juang, C., Chen, C.J., Tien, Y.-M., 1999. Appraising cone penetration test based liquefaction resistance evaluation methods: artificial neural network approach. Can. Geotech. J. 36, 443–454. https://doi.org/10.1139/t99-011

Hsiao, C.-H., Kumar, K., Rathje, E.M., 2024. Explainable AI models for predicting liquefaction-induced lateral spreading. Front. Built Environ. 10, 1387953. https://doi.org/10.3389/fbuil.2024.1387953

Kurup, P.U., Garg, A., 2005. Evaluation of Liquefaction Potential Using Neural Networks Based on Adaptive Resonance Theory. Transportation Research Record: Journal of the Transportation Research Board 1936, 192–200. https://doi.org/10.1177/0361198105193600122

Lai, T., 2024. Interpretable Medical Imagery Diagnosis with Self-Attentive Transformers: A Review of Explainable AI for Health Care. BioMedInformatics.

Li, M., Sun, H., Huang, Y., Chen, H., 2024. Shapley value: from cooperative game to explainable artificial intelligence. Auton. Intell. Syst. 4, 2. https://doi.org/10.1007/s43684-023-00060-8

Liu, F., Liu, W., Li, A., Cheng, J.C.P., 2024. Geotechnical risk modeling using an explainable transfer learning model incorporating physical guidance. Engineering Applications of Artificial Intelligence 137, 109127. https://doi.org/10.1016/j.engappai.2024.109127

Long, T., Akbari, M., Fakharian, P., 2025. Prediction of Soil Liquefaction Using a Multi-Algorithm Technique: Stacking Ensemble Techniques and Bayesian Optimization.

Lundberg, S.M., Lee, S.-I., 2017. A Unified Approach to Interpreting Model Predictions, in: Guyon, I., Luxburg, U.V., Bengio, S., Wallach, H., Fergus, R., Vishwanathan, S., Garnett, R. (Eds.), Advances in Neural Information Processing Systems. Curran Associates, Inc., pp. 4765–4774.

Maurer, B.W., Sanger, M.D., 2023. Why "AI" models for predicting soil liquefaction have been ignored, plus some that shouldn't be. Earthquake Spectra 39, 1883–1910. https://doi.org/10.1177/87552930231173711

National Research Institute for Earth Science and Disaster Resilience, 2019. NIED K-NET, KiK-net. https://doi.org/10.17598/nied.0004

Robertson, P.K., Wride, C.E., 1998. Evaluating Cyclic Liquefaction Potential Using the Cone Penetration Test. Canadian Geotechnical Journal 35, 442–459.

Seed, H.B., Idriss, I.M., 1971. Simplified procedure for evaluating soil liquefaction potential, in: Journal of the Soil Mechanics and Foundations Division. American Society of Civil Engineers, pp. 1249–1273.

USGS, 2025. Vs30 Models and Data [WWW Document]. URL https://earthquake.usgs.gov/data/vs30/ (accessed 2.11.25).

Youd, T.L., Idriss, I.M. (Eds.), 1997. Proceedings of the NCEER Workshop on Evaluation of Liquefaction Resistance of Soils, NCEER-97-0022. National Center for Earthquake Engineering Research (NCEER), Buffalo, NY.
Youwai, S., Detcheewa, S., 2025. Predicting rapid impact compaction of soil using a parallel transformer and long short-term memory architecture for sequential soil profile encoding. Engineering Applications of Artificial Intelligence 139, 109664. https://doi.org/10.1016/j.engappai.2024.109664